\pdfoutput=1

\documentclass[11pt]{article}
\usepackage[preprint]{acl}

\usepackage{times}
\usepackage{latexsym}

\usepackage[T1]{fontenc}

\usepackage[utf8]{inputenc}

\usepackage{microtype}

\usepackage{inconsolata}

\usepackage{graphicx}
\usepackage{hyperref}
\usepackage{xspace}
\usepackage{enumitem}
\usepackage{cleveref}
\usepackage{subcaption}
\usepackage{fancyhdr}
\usepackage{listings,color}
\usepackage[left=1cm,right=1cm,top=1cm,bottom=1.5cm]{}

\usepackage{array}
\usepackage{booktabs}

\title{GRS-QA - Graph Reasoning-Structured Question Answering Dataset}

\author{
\textbf{Anish Pahilajani\textsuperscript{1 *}},
\textbf{Devasha Trivedi\textsuperscript{1 *}},
\textbf{Jincen Shuai\textsuperscript{1 *}},
\textbf{Khin S. Yone\textsuperscript{1 *}},
\\
 \textbf{Samyak Rajesh Jain\textsuperscript{1 *}},
 \textbf{Namyong Park},
 \textbf{Ryan A. Rossi\textsuperscript{2}},
 \textbf{Nesreen K. Ahmed\textsuperscript{3}},
\\
 \textbf{Franck Dernoncourt\textsuperscript{2}},
 \textbf{Yu Wang\textsuperscript{4}}
\\
\\
 \textsuperscript{1}University of California Santa Cruz,
 \textsuperscript{2}Adobe Research,
 \textsuperscript{3}Cisco Outshift,
 \textsuperscript{4}University of Oregon
\\
\\
\texttt{\{apahilaj, detrived, jshuai, kyone, srajeshj\}@ucsc.edu} \hspace{0.3cm} \\
\texttt{park.namyong@gmail.com} \hspace{0.3cm} \texttt{ryrossi@adobe.com} \hspace{0.3cm} \texttt{nesahmed@cisco.com} \hspace{0.3cm} \\ \texttt{dernonco@adobe.com} \hspace{0.3cm} \texttt{yuwang@uoregon.edu} \\
 }

\usepackage{xcolor}
\definecolor{MyForestGreen}{rgb}{0.13, 0.55, 0.13}
\definecolor{MyOrchid}{rgb}{0.85, 0.44, 0.84}
\definecolor{MyRubineRed}{rgb}{0.82, 0.00, 0.34}
\definecolor{MyBurntOrange}{rgb}{0.8, 0.33, 0.0}

\newcommand{\bridgeone}[1]{\textcolor{MyForestGreen}{\textbf{#1}}}
\newcommand{\bridgetwo}[1]{\textcolor{MyOrchid}{\textbf{#1}}}
\newcommand{\bridgethree}[1]{\textcolor{MyRubineRed}{\textbf{#1}}}
\newcommand{\bridgefour}[1]{\textcolor{MyBurntOrange}{\textbf{#1}}}

\begin{document}
\maketitle

\pagestyle{plain} 
\fancypagestyle{firstpage}{ 
  \fancyhf{}
  \fancyfoot[L]{* Equal contribution} 
  \renewcommand{\headrulewidth}{0pt} 
  \renewcommand{\footrulewidth}{0pt} 
}
\thispagestyle{firstpage} 

\begin{abstract}
Large Language Models (LLMs) have excelled in multi-hop question-answering (M-QA) due to their advanced reasoning abilities. However, the impact of the inherent reasoning structures on LLM M-QA performance remains unclear, largely due to the absence of QA datasets that provide fine-grained reasoning structures. To address this gap, we introduce the \underline{G}raph \underline{R}easoning-\underline{S}tructured \underline{Q}uestion \underline{A}nswering Dataset (GRS-QA), which includes both semantic contexts and reasoning structures for QA pairs. Unlike existing M-QA datasets, where different reasoning structures are entangled together, GRS-QA explicitly captures intricate reasoning pathways by constructing reasoning graphs, where nodes represent textual contexts and edges denote logical flows. These reasoning graphs of different structures enable a fine-grained evaluation of LLM reasoning capabilities across various reasoning structures. Our empirical analysis reveals that LLMs perform differently when handling questions with varying reasoning structures. This finding facilitates the exploration of textual structures as compared with semantics.

\end{abstract}

\section{Introduction}

Reasoning in natural language is the fundamental aspect of intelligence~\citep{huang2023reasoninglargelanguagemodels}, and QA tasks provide a quantifiable way to test the reasoning capabilities of intelligent systems\cite{yang2018hotpotqa}. Recently, the emergence of LLMs has demonstrated unprecedented reasoning capacity in answering questions~\cite{wei2022chain, wang2024knowledge}. However, real-world applications often demand even more complex reasoning capability, such as multi-hop reasoning~\cite{10.1145/3539618.3591698}, where systems must integrate information from multiple sources and perform multiple steps of thinking in a certain order to arrive at the final answer and conclusion.

To evaluate the multi-hop reasoning capabilities of LLMs, researchers have developed several multi-hop question-answering (M-QA) datasets, including HotpotQA, 2WikiMultiHopQA, and MuSiQue~\cite{yang2018hotpotqa, ho2020constructing, trivedi2022musique}. MuSiQue constructs genuine multi-step QA pairs by composing connected single-hop questions through a bottom-up approach and mitigating existing common shortcuts. 2WikiMultiHopQA integrates structured and unstructured data to provide comprehensive and evidence-based reasoning paths, which ensures authentic multi-hop reasoning. HotpotQA is a large-scale and crowd-sourced dataset comprising 113,000 Wikipedia-based QA pairs, offering sentence-level supporting facts for explainable predictions and introducing challenging comparison questions. Despite their contributions to benchmarking LLMs' multi-hop reasoning capabilities, these datasets have notable limitations. First, most current M-QA datasets lack explicit reasoning structures for each QA pair, preventing LLMs from leveraging predefined reasoning pathways and forcing them to rely solely on their internal knowledge. Second, they mix questions with varying reasoning complexities without categorization, making it challenging to investigate LLMs' QA capability at a fine-grained structure level of the question.

To address these limitations, we introduce GRS-QA, a novel \underline{G}raph \underline{R}easoning-\underline{S}tructured \underline{Q}uestion \underline{A}nswering Dataset, that integrates explicit reasoning structure in the format of the graph (i.e., reasoning graph) to enable multi-hop reasoning analysis of LLM-based MQA. By incorporating reasoning graphs that represent multi-hop inference pathways, GRS-QA offers several advantages. First, these graphs provide a transparent way to understand the logical steps LLMs should follow to reach the answer, making the reasoning process explicit and allowing researchers to pinpoint where a model may struggle at a fine-grained level. Second, the reasoning graphs are categorized based on their structural complexity. Furthermore, each reasoning graph is accompanied by metadata (e.g., the number of reasoning steps and types), which facilitates analysis to identify patterns in question difficulty, reasoning complexity, and LLM performance. Thirdly, the structured nature of the reasoning graphs enables the development of new evaluation metrics, such as reasoning recall/precision, that go beyond answer correctness and allow the assessment of LLMs to replicate the reasoning pathway.

To construct the reasoning graph, we treat each sentence as a node and add edges based on their original logical relations. In addition, to further investigate the importance of the reasoning structure in correctly answering the question compared to the content, we also generate structural negative samples by adding noise to their graph structure, such as adding extra sentence nodes and rewiring the reasoning graphs. Investigating the QA performance when paired with these additionally introduced negative samples helps demystify the importance of the structure in QA. Our contribution can be summarized as follows:

\begin{itemize}[leftmargin=*]
    \item \textbf{First QA Dataset with Reasoning Graphs}: GRS-QA introduces the first question-answering dataset that pairs each instance with explicit reasoning graphs. These reasoning graphs provide a transparent pathway to understand the logical steps that LLMs should follow to reach the correct answer, allowing researchers to pinpoint specific areas where a model may struggle.

    \item \textbf{Comprehensive Analysis and Categorization of Reasoning Graphs}: Each QA pair in GRS-QA is accompanied by comprehensive metadata, such as reasoning types and reasoning complexity. This enables performance analysis according to the reasoning structure and offers new insights into how LLMs perform on questions with different levels of reasoning complexity.

    \item \textbf{Negative Reasoning Graphs}: In addition to the ground-truth reasoning graph, each QA pair also includes corresponding negative reasoning graphs where the content remains the same, but the structure is slightly modified. This enables an exclusive analysis of the structural impact on reasoning and QA performance.

\end{itemize}

\section{GRS-QA Dataset Construction}

\begin{figure*}[htbp]
    \centering
    \includegraphics[width=1\linewidth]{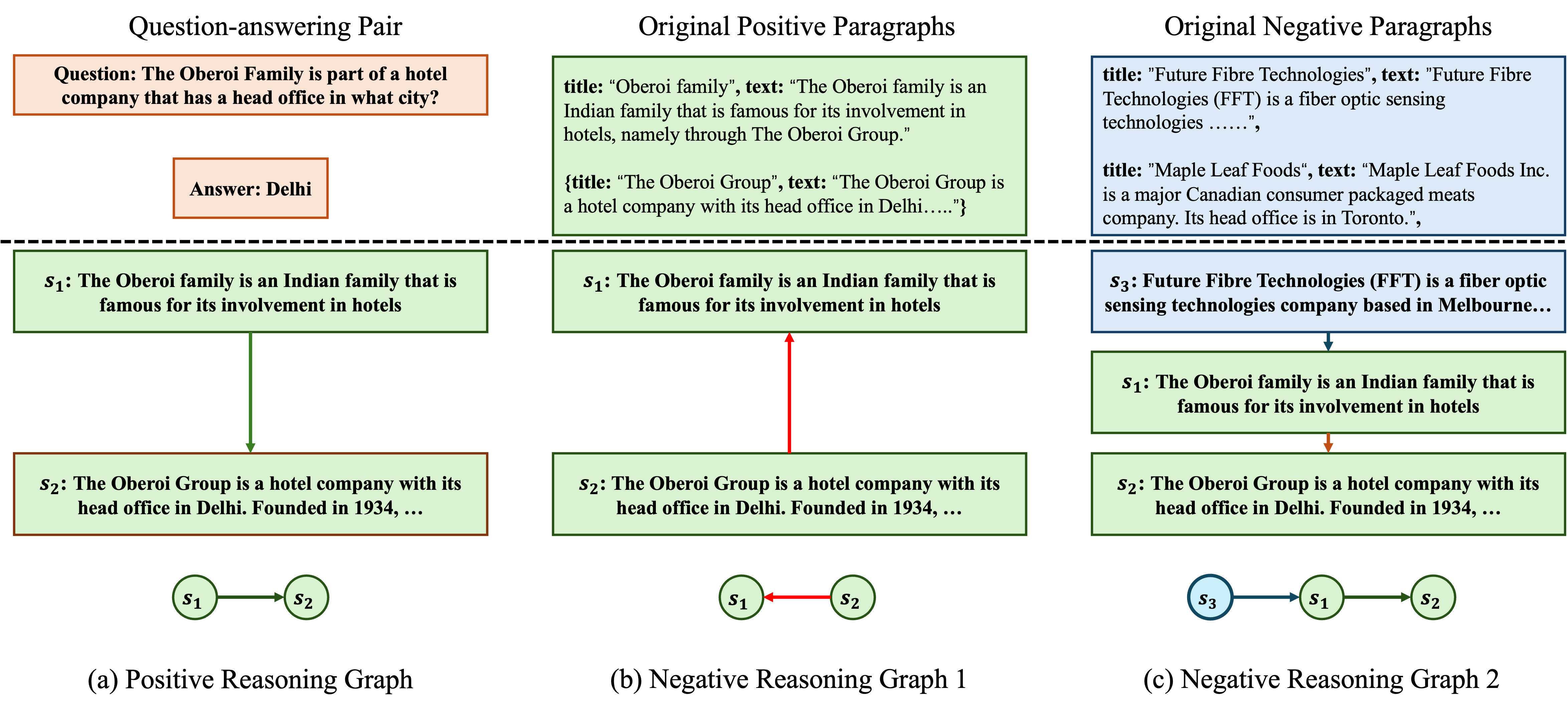}
    \caption{
    Reasoning graphs constructed based on one QA instance from HotpotQA dataset~\cite{yang2018hotpotqa} that maps out the logical steps required to arrive at the answer. The left-hand side illustrates the \textbf{positive reasoning graph}, which is constructed from the supporting paragraphs provided in the original dataset. This graph represents the \textit{gold reasoning path} needed to answer the question. On the right-hand side, two types of \textbf{negative reasoning graphs} are derived from the original positive reasoning graphs by either perturbing the edges (e.g., inversing the edge direction in this case) or adding additional nodes with irrelevant sentences.
    }
    \label{fig:hotpot-processing}
\end{figure*}

As motivated before, our work introduces a new dataset, GRS-QA, which contains reasoning graphs that trace the logical paths from questions to answers. Specifically, we construct these graphs using QA pairs in the training sets of three existing multi-hop QA datasets: HotpotQA, MuSiQue, and 2WikiMultiHopQA, which provide diverse and comprehensive reasoning structures. These datasets have uniform fields, including a question, answer, and supporting facts that serve as the golden contexts for answering the question. To build the reasoning graph, we treat each sentence as a node and add edges denoting their connections based on the local relations derived from the original dataset. \Cref{fig:hotpot-processing} shows an example of building the positive and negative reasoning graphs for one QA instance from HotpotQA. Next, we will introduce the details of building GRS-QA.

\subsection{Positive Reasoning Graphs}
Positive reasoning graphs illustrate the golden reasoning steps from a question to its answer. In these graphs, nodes represent sentences of the golden context that address portions of the question. Edges between nodes define the logical flow that the LLM should follow to arrive at the correct answer.

In the HotpotQA dataset, questions are categorized as either "bridge" or "comparison." Each question is paired with two sentences from the golden context, which serve as nodes in a reasoning graph. For "comparison" questions, no edges are established between nodes. For "bridge" questions, we use the "keyword" field to identify the second sentence as the tail node, designate the other sentence as the head node, and create an edge from head to tail. In the MuSiQue dataset, we utilize the provided question category (e.g., "4hop2") to determine the graph structure. The supplied sentence IDs are used to set up the nodes accordingly. For 2WikiMultiHopQA, the triplets in the "evidences" field provides sufficient information of establishing the edges for our graph. We extract entities from the initial sentence to locate their corresponding sentences. Additionally, the MuSiQue dataset includes an `answerable' field that indicates whether the provided context contains the necessary sentences to address parts of the question. If the "answerable" field is marked as False, we exclude that data point from GRS-QA, as we cannot generate a reasoning graph with incomplete context.

\subsection{Negative Reasoning Graphs}

In addition to the golden reasoning graphs that map out the ground-truth logical steps to answer each question, we also create negative reasoning graphs by perturbing the structure of the positive reasoning graphs. Our goal here is to exclusively investigate the impact of the structure of the logic flow rather than the content of the sentences.

Unlike conventional negative samples, where the context differs semantically from the golden one (e.g., the negative samples could be the ones with totally/slightly different sentences from the ground truth), we focus on structural differences. Specifically, we consider two types of structural perturbations: edge perturbation and node perturbation. For edge perturbation, we keep the original nodes but alter the edges by addition, removal, and rewriting, thereby changing the reasoning flow. For node perturbation, we randomly remove some nodes, add new nodes, or swap out nodes with others. The nodes we swap out or add are sourced from the global sentence pool except the ones of the positive graphs. Taking the positive reasoning graph in Figure 1(a) as an example, we reverse the edge from sentence $s_1$ to $s_2$ and create the negative reasoning graph 1 with the incorrect logic flow. Meanwhile, we add another irrelevant sentence $s_3$ as node 3 and add an edge between it and node $s_1$ to create the negative reasoning graph 2.

\begin{figure*}[t!] 
    \centering
    \begin{subfigure}[b]{0.32\linewidth}
        \centering
        \includegraphics[width=1\linewidth]{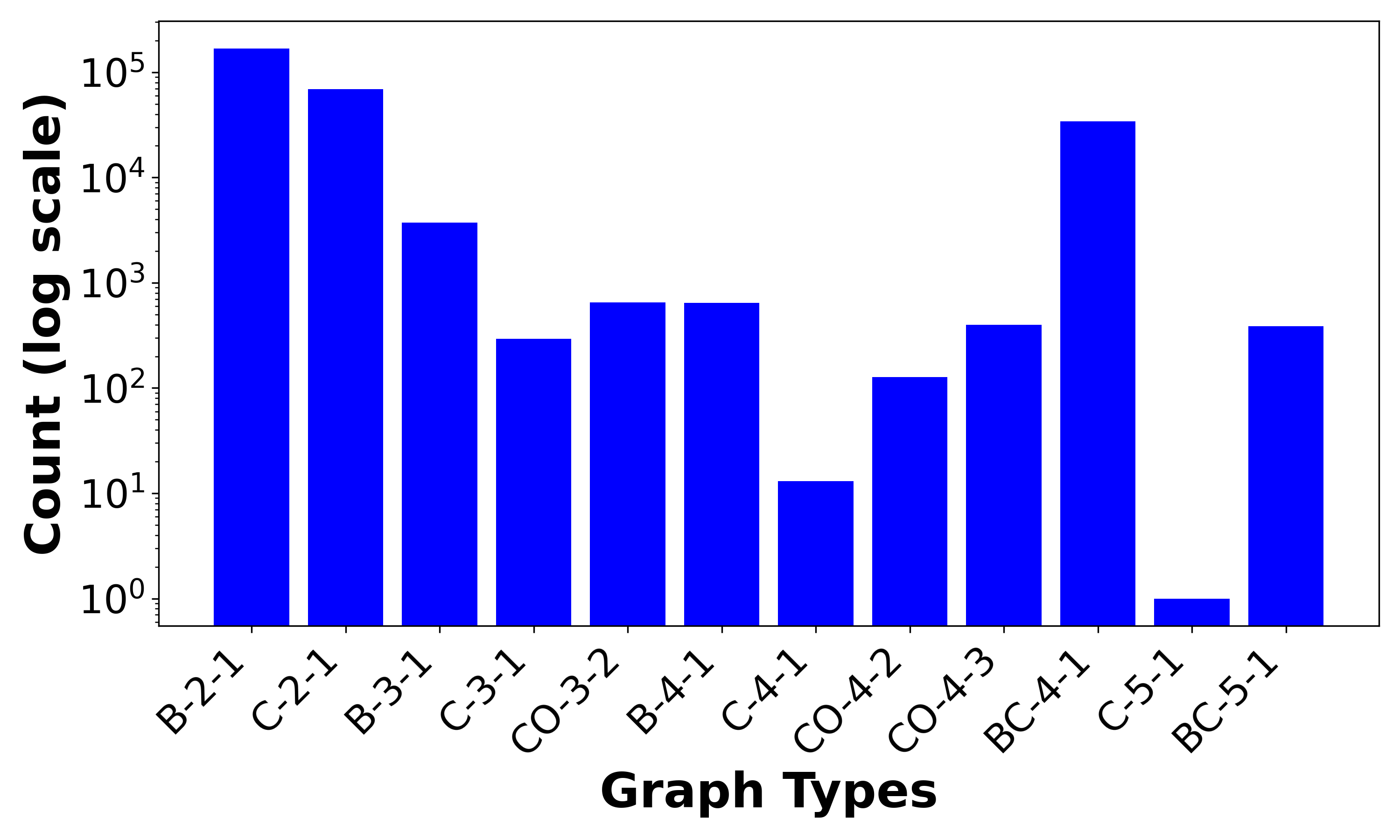}
        \caption{Number of Questions by Graph types in all dataset splits} 
        \label{fig:percentages-graphs}
    \end{subfigure}
    \hfill
    \begin{subfigure}[b]{0.32\linewidth}
        \centering
        \includegraphics[width=1\linewidth]{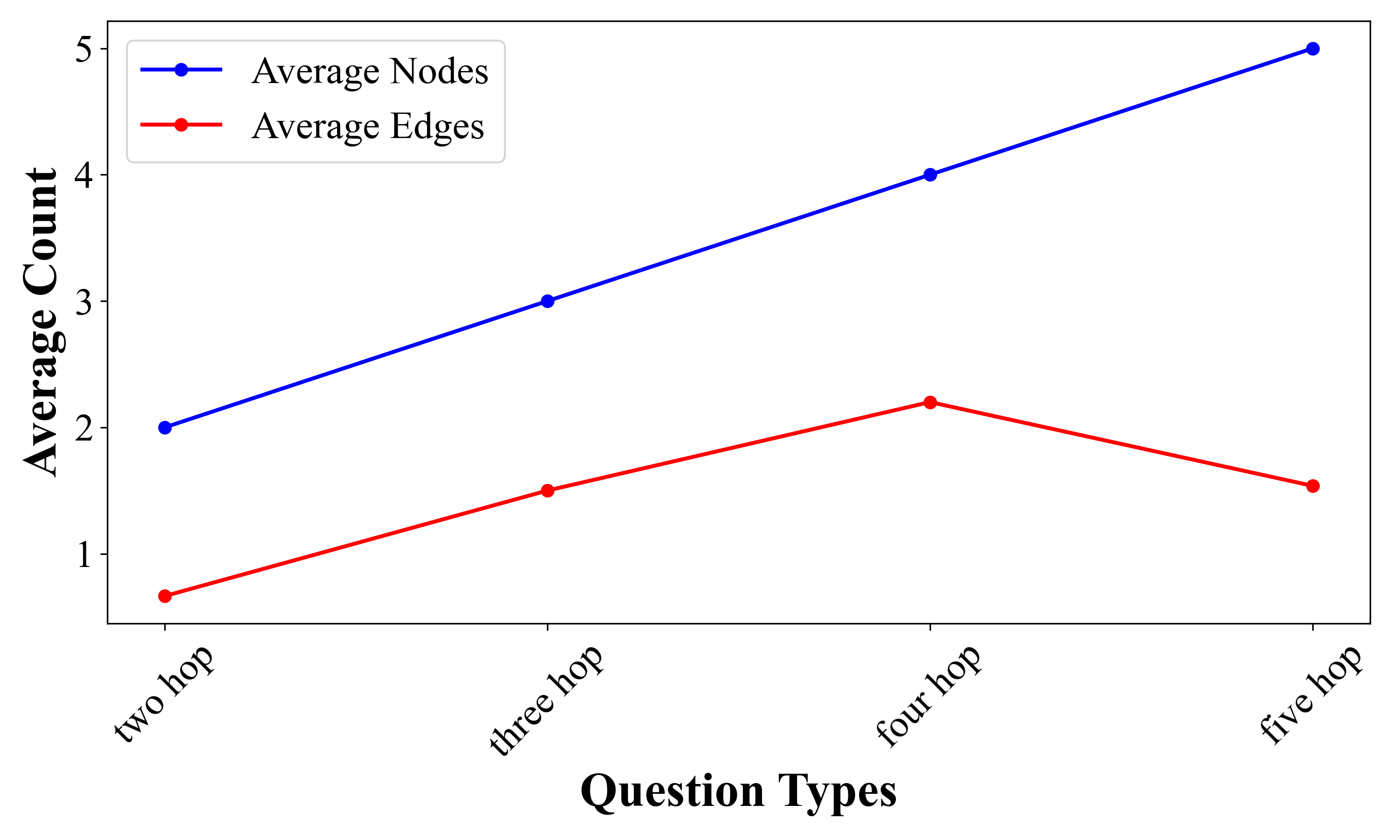}
        \caption{Average number of nodes and edges in each question type Positive Graphs}
        \label{fig:avg-edge-count}
    \end{subfigure}
    \hfill
    \begin{subfigure}[b]{0.32\linewidth}
        \centering
        \includegraphics[width=1\linewidth]{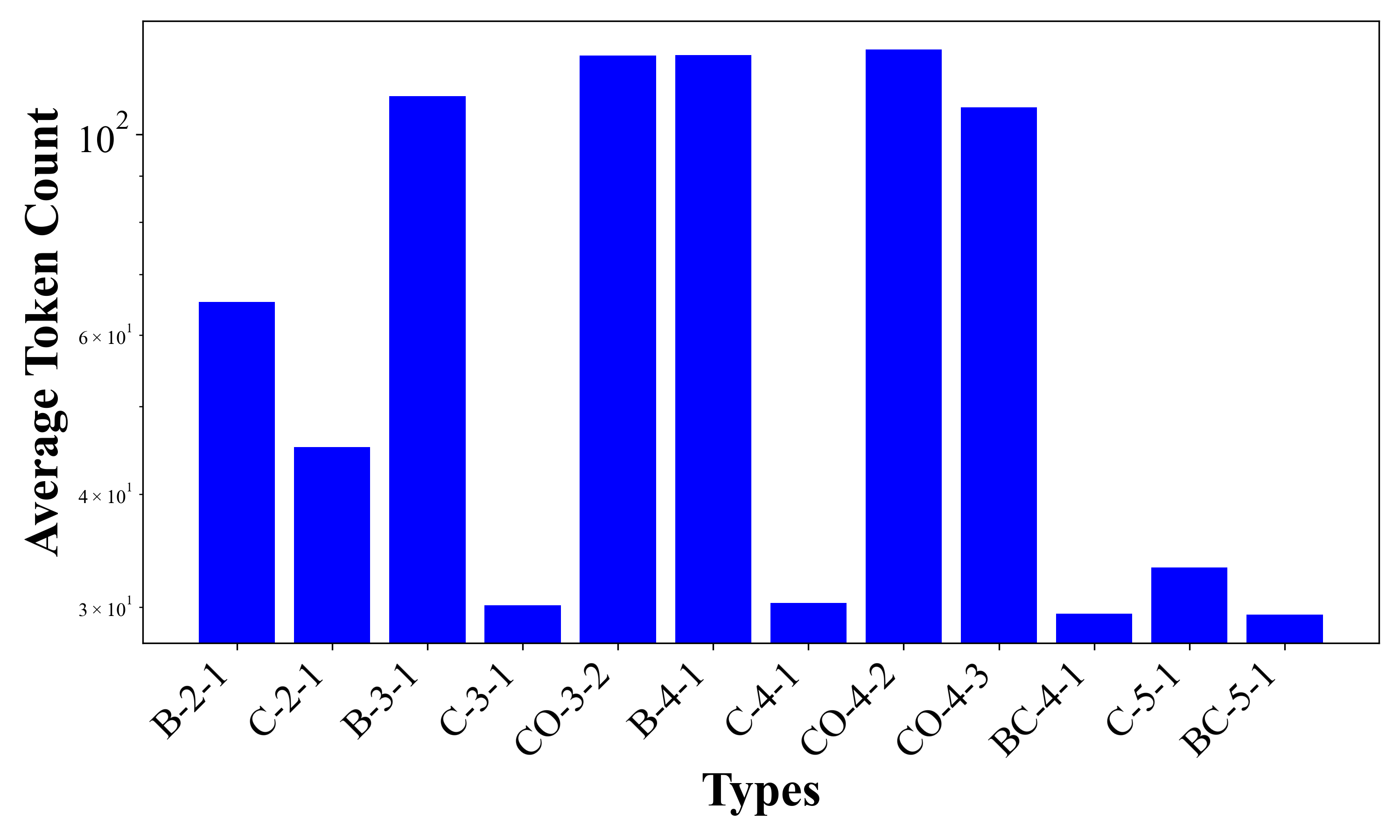}
        \caption{Average number of tokens in each question type's Positive Graphs}
        \label{fig:average_token_count}
    \end{subfigure}
    \caption{Statistical Analysis of the Distribution of GRS-QA.}
    \label{fig:dataset_dist}
\end{figure*}

\begin{table*}[htbp]
    \scriptsize
    \centering
    \resizebox{\textwidth}{!}{
    \renewcommand{\arraystretch}{0}
    \begin{tabular}{
        >{\centering\arraybackslash}p{2cm}
        p{2.5cm}p{3.8cm}p{8cm}
    }

    \toprule
        
        Graph & Type & Question & Decomposition\\
        \midrule
        \raisebox{-0.8\height}{\includegraphics[height=0.33\linewidth]{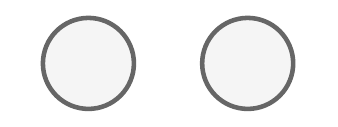}} & Comparison\_2\_1 (C-2-1) &
        Between Athlete and Fun, which band has more members? \bridgetwo{Athlete} & 

        \textbf{1.} How many members are in Athlete? \bridgeone{Four members} \newline
        \textbf{2.} How many members are in Fun? \bridgetwo{Three members}
        \\ 
        \midrule
        
        \raisebox{-0.8\height}{\includegraphics[height=0.25\linewidth]{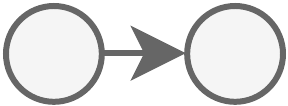}} & Bridge\_2\_1 (B-2-1) &
        Who beat the player that won the 2017 Australian men's open tennis single title in the US open? \bridgetwo{Novak Djokovic} &

        \textbf{1.} Who wins the 2017 australian men's open tennis single title? \bridgeone{Roger Federer} \newline
        \textbf{2.} Who beat \bridgeone{Roger Federer} in the us open? \bridgetwo{Novak Djokovic}
        \\
        \midrule
        
        \raisebox{-0.8\height}{\includegraphics[height=0.8\linewidth]{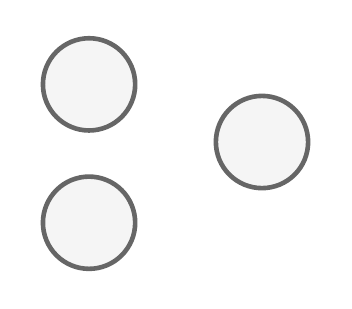}} & Comparison\_3\_1 (C-3-1) &
        In which country is the administrative territorial entity for the city where Charlie Harper was born? \bridgethree{United Kingdom} &

        \textbf{1.} Where was Charlie Harper born? \bridgeone{Hackney} \newline
        \textbf{2.} In which administrative territorial entity is \bridgeone{Hackney} located? \bridgetwo{Middlesex} \newline
        \textbf{3.} Which country is \bridgetwo{Middlesex} located in? \bridgethree{United Kingdom}
        \\
        \midrule
        
        \raisebox{-0.8\height}{\includegraphics[height=0.25\linewidth]{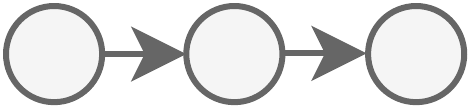}} & Bridge\_3\_1 (B-3-1) &
        In which country is the administrative territorial entity for the city where Charlie Harper was born? \bridgethree{United Kingdom} &

        \textbf{1.} Where was Charlie Harper born? \bridgeone{Hackney} \newline
        \textbf{2.} In which administrative territorial entity is \bridgeone{Hackney} located? \bridgetwo{Middlesex} \newline
        \textbf{3.} Which country is \bridgetwo{Middlesex} located in? \bridgethree{United Kingdom}
        \\
        \midrule
       
        \raisebox{-0.8\height}{\includegraphics[height=0.6\linewidth]{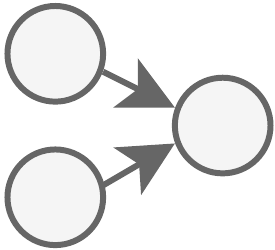}} & Compositional\_3\_2 (CO-3-2) &
        In which country is Midway, in the same county as McRae in the same state as KAGH-FM? \bridgethree{U.S.} &

        \textbf{1.} What state is KAGH-FM located? \bridgeone{Arkansas} \newline
        \textbf{2.} In which administrative territorial entity is McRae located? \bridgetwo{White County} \newline
        \textbf{3.} Which country is Midway (near Pleasant Plains), \bridgetwo{White County}, \bridgeone{Arkansas} located in? \bridgethree{U.S.} 
        \\
        \midrule
        
        \raisebox{-0.8\height}{\includegraphics[height=1.0\linewidth]{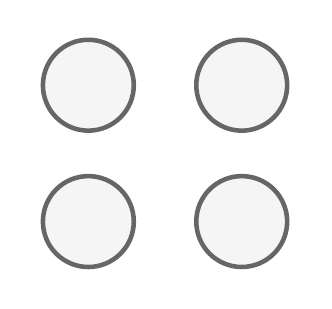}} & Comparison\_4\_1 (C-4-1) &
        Did Albrecht Alt and Asli Hassan Abade have the same occupation? \bridgefour{no} &

        \textbf{1.} ["Asli Hassan Abade", "occupation", "pilot"]\newline
        \textbf{2.} ["Asli Hassan Abade", "occupation", "military figure"], \newline
        \textbf{3.} ["Asli Hassan Abade", "occupation", "civil activist"] \newline
        \textbf{4.} ["Albrecht Alt", "occupation", "theologian"] \newline
        \textbf{5.} ["Albrecht Alt", "occupation", "lecturer"]\newline
        \textbf{6.} ["Albrecht Alt", "occupation", "professor"] \newline
        "supporting\_facts": [["Asli Hassan Abade", 0], ["Albrecht Alt", 0],["Albrecht Alt", 2], ["Albrecht Alt", 6]]
        \\
        \midrule
        
        \raisebox{-0.9\height}{\includegraphics[height=0.75\linewidth]{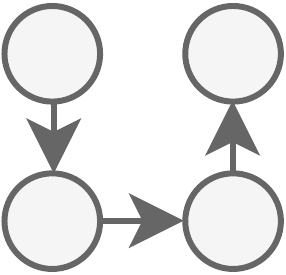}} & Bridge\_4\_1 (B-4-1) &
        When did Ukraine gain independence from the first Allied nation to reach the German city where the director of The Man from Morocco was born? \bridgefour{1917} &

        \textbf{1.} Who is the director of The Man from Morocco? \bridgeone{Mutz Greenbaum} \newline
        \textbf{2.} What is the place of birth of \bridgeone{Mutz Greenbaum}? \bridgetwo{Berlin} \newline
        \textbf{3.} What allied nation was the first to reach the german capitol of \bridgetwo{Berlin}? \bridgethree{Soviet Union} \newline
        \textbf{4.} When did Ukraine gain independence from \bridgethree{Soviet Union}? \bridgefour{1917}
        \\
        \midrule
        
        \raisebox{-1\height}{\includegraphics[height=0.6\linewidth]{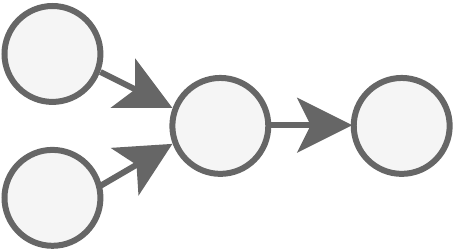}} & Compositional\_4\_2 (CO-4-2) &
        Where is the place of death of the man who became leader of the largest country in Europe in square miles after the collapse of the nation Germany agreed to sign a non-aggression pact with in 1939? \bridgefour{Moscow} &

        \textbf{1.} What is the largest country in europe by square miles? \bridgeone{Russia} \newline
        \textbf{2.} In 1939 Germany agreed to sign a non-aggression pact with which country? \bridgetwo{the Soviet Union} \newline
        \textbf{3.} Who became leader of \bridgeone{Russia} after the collapse of \bridgetwo{the Soviet Union}? \bridgethree{Boris Yeltsin} \newline
        \textbf{4.} Where did \bridgethree{Boris Yeltsin} die? \bridgefour{Moscow}
        \\
        \midrule
       
        \raisebox{-1\height}{\includegraphics[height=0.6\linewidth]{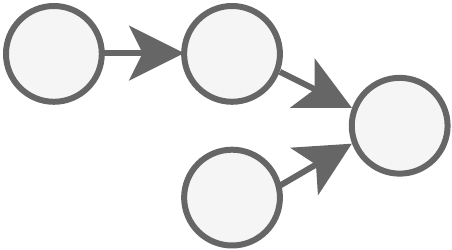}} & Compositional\_4\_3 (CO-4-3) &
        In what country is Tuolumne, which is within a county that borders the county containing Jamestown, and is located within the state where Some Like It Hot was filmed? \bridgefour{United States} &

        \textbf{1.} In which administrative territorial entity is Jamestown located? \bridgeone{Tuolumne County} \newline
        \textbf{2.} Which entities share a border with \bridgeone{Tuolumne County}? \bridgetwo{Stanislaus County} \newline
        \textbf{3.} Where did they film some like it hot? \bridgethree{in California} \newline
        \textbf{4.} Which country is \bridgeone{Tuolumne}, \bridgetwo{Stanislaus County}, in California located in?? \bridgefour{United States} 
        \\
        \midrule
        
        \raisebox{-0.8\height}{\includegraphics[height=1.0\linewidth]{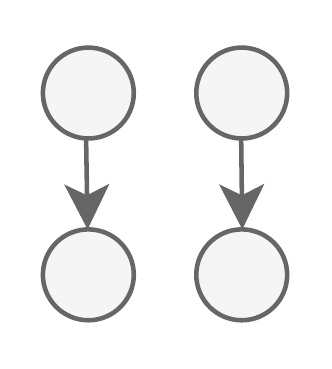}} & Bridge\_Comparison\_4\_1 (BC-4-1) &
        Are both directors of films The Blue Bird (1940 Film) and Bharya Biddalu from the same country? \bridgefour{no} &

        \textbf{1.} ['The Blue Bird (1940 film)', 'director', \bridgeone{'Walter Lang'}]  \newline
        \textbf{2.} ['Bharya Biddalu', 'director', \bridgetwo{'Tatineni Rama Rao'}]  \newline
        \textbf{3.} [\bridgeone{'Walter Lang'}, 'country of citizenship', 'American']  \newline
        \textbf{4.} [\bridgetwo{'Tatineni Rama Rao'}, 'country of citizenship', 'India'] 
        \\
        \midrule
        
         \raisebox{-0.8\height}{\includegraphics[height=0.9\linewidth]{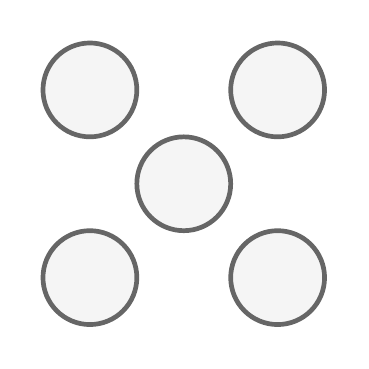}} & Comparison\_5\_1 (CO-5-1) &
         Which film has more directors, Red Cow (Film) or Chillerama? \bridgefour{Chillerama} &

        \textbf{1.} ["Red Cow (film)", "director", "Tsivia Barkai Yacov"]  \newline
        \textbf{2.} ["Chillerama", "director", "Adam Rifkin"]  \newline
        \textbf{3.} ["Chillerama", "director", "Tim Sullivan"] \newline
        \textbf{4.} ["Chillerama", "director", "Adam Green"]\newline
        \textbf{5.} ["Chillerama", "director", "Joe Lynch"]
        \\
        \midrule
        
         \raisebox{-0.8\height}{\includegraphics[height=0.85\linewidth]{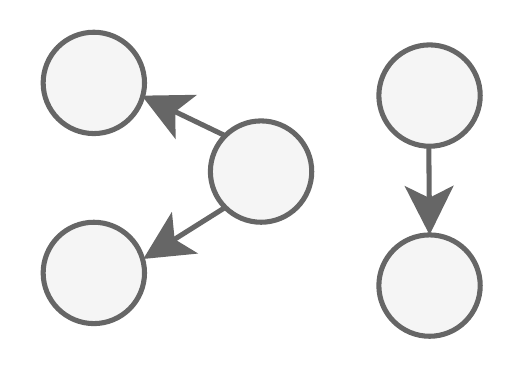}} & Bridge\_Comparison\_5\_1 (BC-5-1) &
        "Do both films The Falcon (Film) and Valentin The Good have the directors from the same country? \bridgefour{no} &

        \textbf{1.}  ["The Falcon (film)", "director", \bridgeone{"Vatroslav Mimica"}]  \newline
        \textbf{2.} ["Valentin the Good", "director", \bridgetwo{"Martin Fri\u010d"}]  \newline
        \textbf{3.} [\bridgeone{"Vatroslav Mimica"}, "country of citizenship", "Croatian"]  \newline
        \textbf{4.} [\bridgeone{"Vatroslav Mimica"}, "country of citizenship", "Yugoslavia"] \newline
        \textbf{5.} [\bridgetwo{"Martin Fri\u010d"}, "country of citizenship", "Czech"]
        
        \\

        \bottomrule
    \end{tabular}
    }

    \caption{
        This table shows the Reasoning graphs of GRS-QA. The reasoning graphs demonstrate the decomposition of the larger question and the reasoning paths to approach the answer. Each of these is constructed using the context and relevant entities for each question. The decomposition is shown with varying formats in the right-most column of the graph, including more questions derived from the original question as well as triples that represent the relations between entities and, in turn, provide subsets of the context. This is consistent with the multiple datasets that each of the question types are extracted from. 
    }

    \label{table:reasoning-graphs}
\end{table*}

\subsection{Dataset Distribution}
The constructed reasoning graphs are categorized into four main types based on their logical structure: comparison, bridge, compositional, and bridge-comparison, resulting in 12 unique graph structures. Comparison graphs lack edge connections, while bridge graphs feature sequential edges. Bridge-comparison graphs consist of two disconnected components, each resembling a bridge. Compositional graphs are tree-structured, where some branches may form bridges. Table~\ref{table:reasoning-graphs} outlines the profiles of the QA pairs and their corresponding reasoning graphs. Additionally, the statistics of the GRS-QA dataset are illustrated in \Cref{fig:dataset_dist}, showing the distribution of question types (a), the number of hops (b), and the average token count (c). The majority of the dataset comprises bridge and comparison questions, aligning with their frequency in the underlying datasets and their common use in everyday reasoning. \Cref{fig:dataset_dist} shows the average number of nodes and edges for questions with 2 to 5 hops. There is a clear proportionality between edge counts and node numbers for 2-hop to 4-hop questions, but 5-hop questions deviate due to their unique graph structures. Lastly, \Cref{fig:dataset_dist} displays the average token count for each graph type, revealing that graphs with simpler structures tend to have longer sentences, leading to a higher token count.

\section{Experiments}

To assess how effectively GRS-QA challenges state-of-the-art models, we benchmark their performance from three key perspectives. First, we evaluate how well retrievers can fetch the correct sentences that form the positive reasoning graphs from a pool of candidate sentences. Second, we assess LLM performance on directly answering questions with different reasoning structures without providing any supporting context. Finally, we measure the combined retrieval and generation performance, where LLMs generate answers using various forms of retrieved evidence.

To ensure consistency and comparability, we use the same question set across all experiments. To balance statistical significance with budget constraints, we capped the number of questions per type at 100. For question types with fewer than 100 entries in the test set, we used all available questions. In total, we tested 660 questions across 12 different question types. The experiments are all conducted on a shared GPU server equipped with six NVIDIA RTX 3090 devices with a total of 10 GPU Hours.

\subsection{Retrieval Performance Benchmark}
To evaluate retrieval performance on our GRS-QA dataset, we employ three methods: Best Matching 25 (BM25)~\cite{robertson2009probabilistic}, Dense Passage Retrieval (DPR)~\cite{karpukhin2020dense}, and Term Frequency-Inverse Document Frequency (TF-IDF). The objective is to determine whether retrieving relevant evidence sentences becomes more challenging with increasingly complex reasoning structures. For BM25, we use BM25Okapi from the built-in Python library. For DPR, we utilize the pretrained model \href{https://huggingface.co/facebook/dpr-question_encoder-single-nq-base}{from Facebook}. For TF-IDF, we use the Vectorizer provided by scikit-learn. The retrieval evaluation process is as follows. First, we pull different question types from the test split of GRS-QA and, for each question, compare it against a bag of all unique evidence sentences in the test set. Using one of the three retrieval methods (BM25, DPR, or TF-IDF), we retrieve the top 20 evidence sentences. We then calculate F1, precision, and recall by comparing the retrieved sentences with the ground truth sentences.

Our results show a general trend: BM25 performance drops as the questions become more complex (e.g., multi-hop questions), while DPR and TF-IDF display inconsistent behavior across different question types. In some cases, their retrieval metrics are nearly zero. Despite these outliers, BM25 outperforms both DPR and TF-IDF on average, as shown in \autoref{fig:bm25_retrieval}, \autoref{fig:dpr_retrieval}, and \autoref{fig:tfidf_retrieval}. We conclude that BM25 is our best retriever model, and the poor performance of DPR on average can be attributed to the fact that we do not train a DPR model using the train split of our data. The comprehensive average performance of each retriever on all question types is presented in \autoref{tab:retrieval_metrics} in the appendix.

\begin{figure*}[t!]
    \centering
    \begin{subfigure}[b]{0.49\linewidth}
        \centering
        \includegraphics[width=\linewidth]{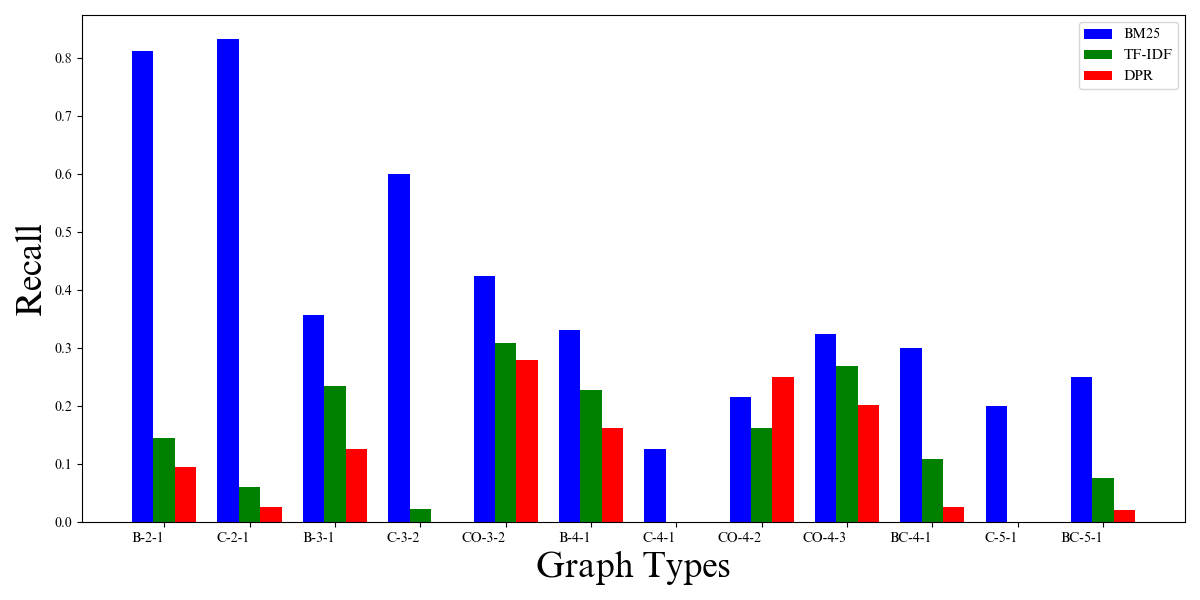}
        \caption{Recall Across Question of Different Reasoning Graphs}
        \label{fig:recall-comparison}
    \end{subfigure}
    \hfill
    \begin{subfigure}[b]{0.49\linewidth}
        \centering
        \includegraphics[width=\linewidth]{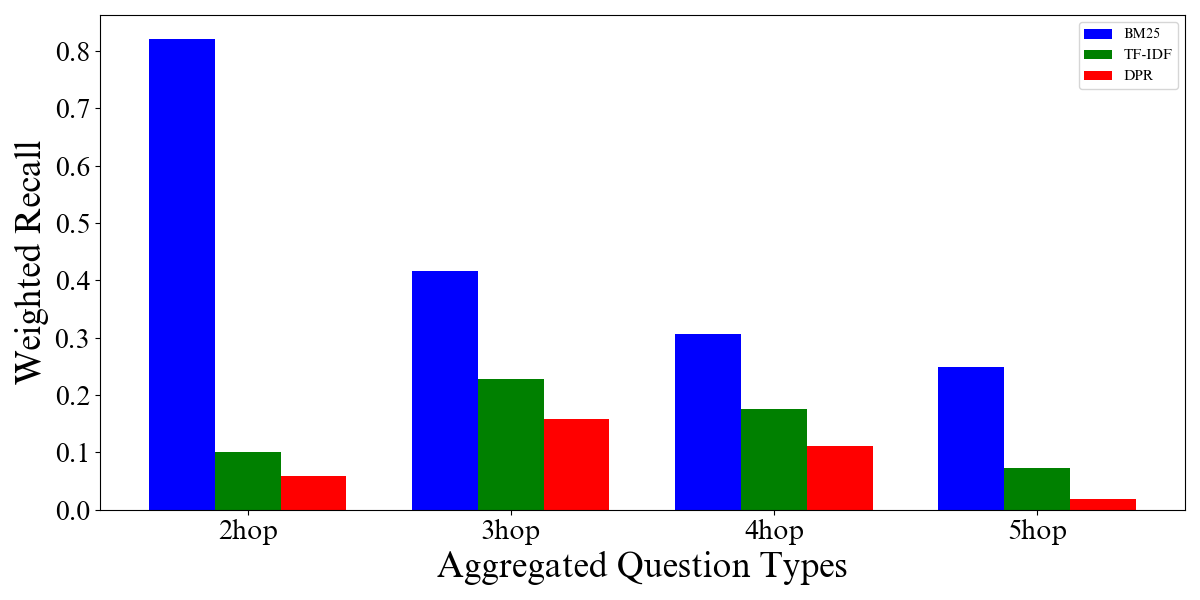}
        \caption{Weighted Recall Across Questions of Different Hops}
        \label{fig:weighted-recall}
    \end{subfigure}
    \caption{Comparison of BM25, TFIDF, and DPR Recall and Weighted Recall Across Question Types}
    \label{fig:combined-figures}
\end{figure*}

\subsection{LLM QA Performance Benchmark}

To evaluate the LLM QA performance, we select three LLMs: Llama3(8B Instruct)~\cite{dubey2024llama3herdmodels}, GPT-3.5~\cite{openai_gpt3_5_turbo}, and GPT4o-mini~\cite{openai_gpt4o_mini}. Their performance is measured using three metrics: exact match, F1 score, and LLM-as-Judge~\cite{zheng2023judgingllmasajudgemtbenchchatbot}, with GPT4o-mini used as the judge. For all experiments, we use a temperature of 0.2 for the LLMs and employ float16 precision for the Llama3 8B model. Our results indicate that GPT-3.5 achieves the highest performance, while Llama3 shows comparatively lower performance. \autoref{fig:3.5_direct}, \autoref{fig:4o-mini_direct} and \autoref{fig:llama3_direct} display the performance of all three LLMs across questions of different reasoning graph complexity. On average, as question complexity increases, LLM performance decreases, though there are exceptions like the comparison\_5 question type. This is likely due to only one comparison\_5 question in that specific set. Additionally, we had only around 41 5-hop questions compared to approximately 200 each of 2-hop, 3-hop, and 4-hop questions. This scarcity affects the statistical significance of the observed performance. Despite this, the observed trend indeed highlights that the underlying reasoning structure closely affects the M-QA performance.

\begin{figure*}[htbp]
    \centering
    \begin{subfigure}{0.32\textwidth}
        \centering
        \includegraphics[width=\linewidth]{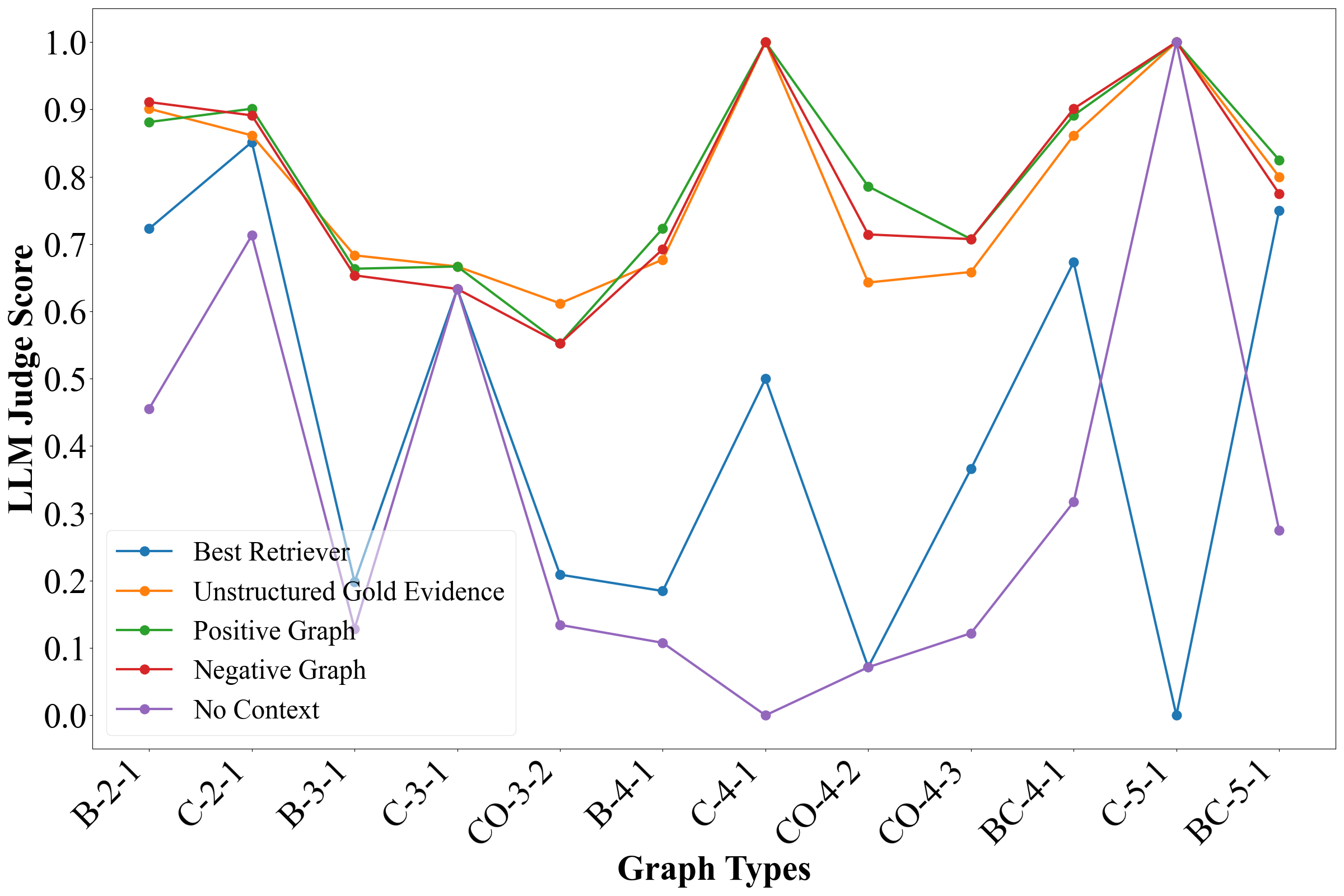}
        \caption{GPT-3.5 as LLM-Judge}
        \label{fig:3.5_all_prompts}
    \end{subfigure}
    \hfill
    \begin{subfigure}{0.32\textwidth}
        \centering
        \includegraphics[width=\linewidth]{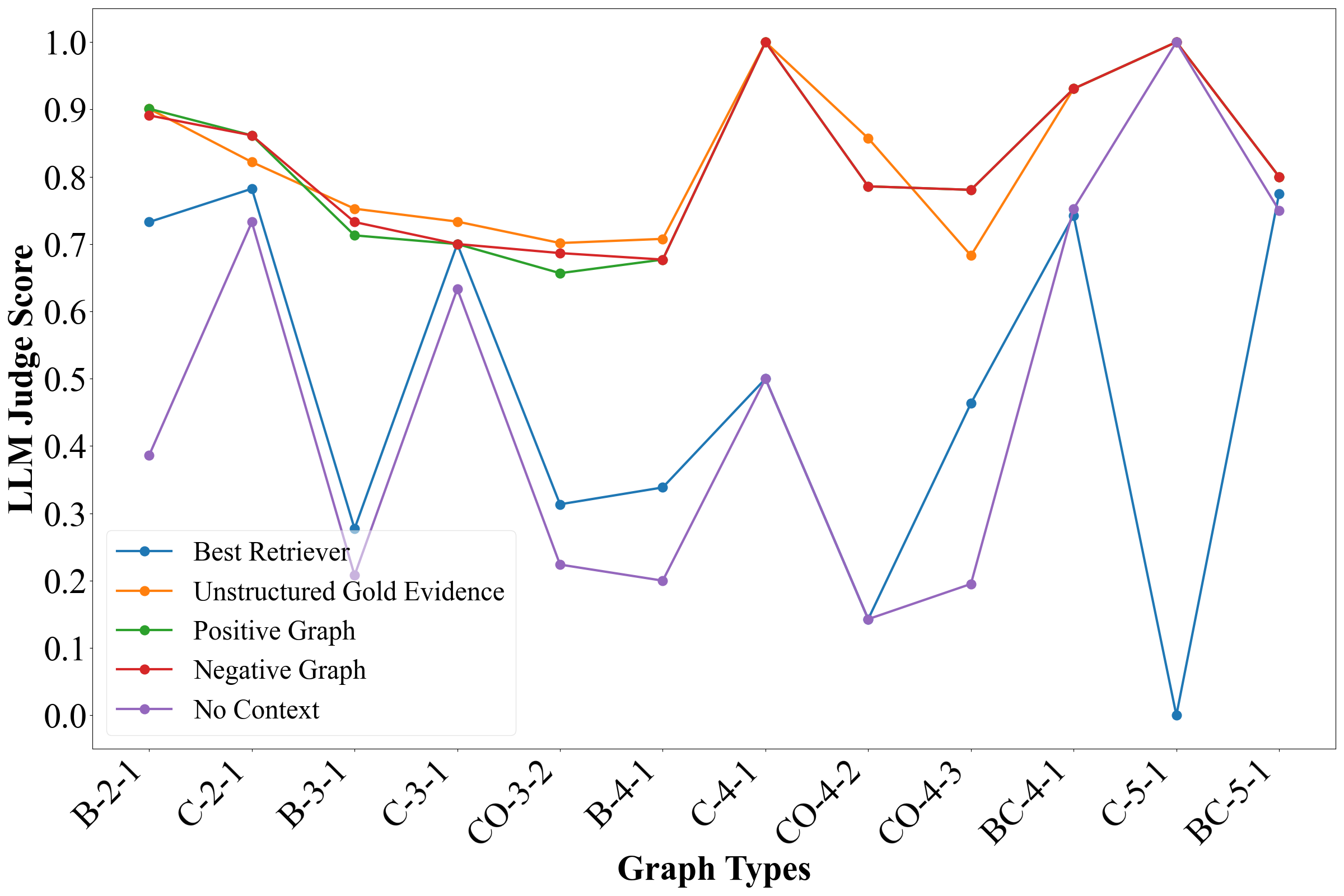}
        \caption{GPT-4o-mini as LLM-Judge}
        \label{fig:4o_all_prompts}
    \end{subfigure}
    \hfill
    \begin{subfigure}{0.32\textwidth}
        \centering
        \includegraphics[width=\linewidth]{images/question_types_GPT-3.5.png}
        \caption{Llama3 as LLM-Judge}
        \label{fig:llama_all_prompts}
    \end{subfigure}
    \caption{LLM Judge Scores by Question Type for Different LLMs}
    \label{fig:question_types}
\end{figure*}

\begin{figure*}[htbp]
    \centering
    \begin{subfigure}{0.32\textwidth}
        \centering
        \includegraphics[width=\linewidth]{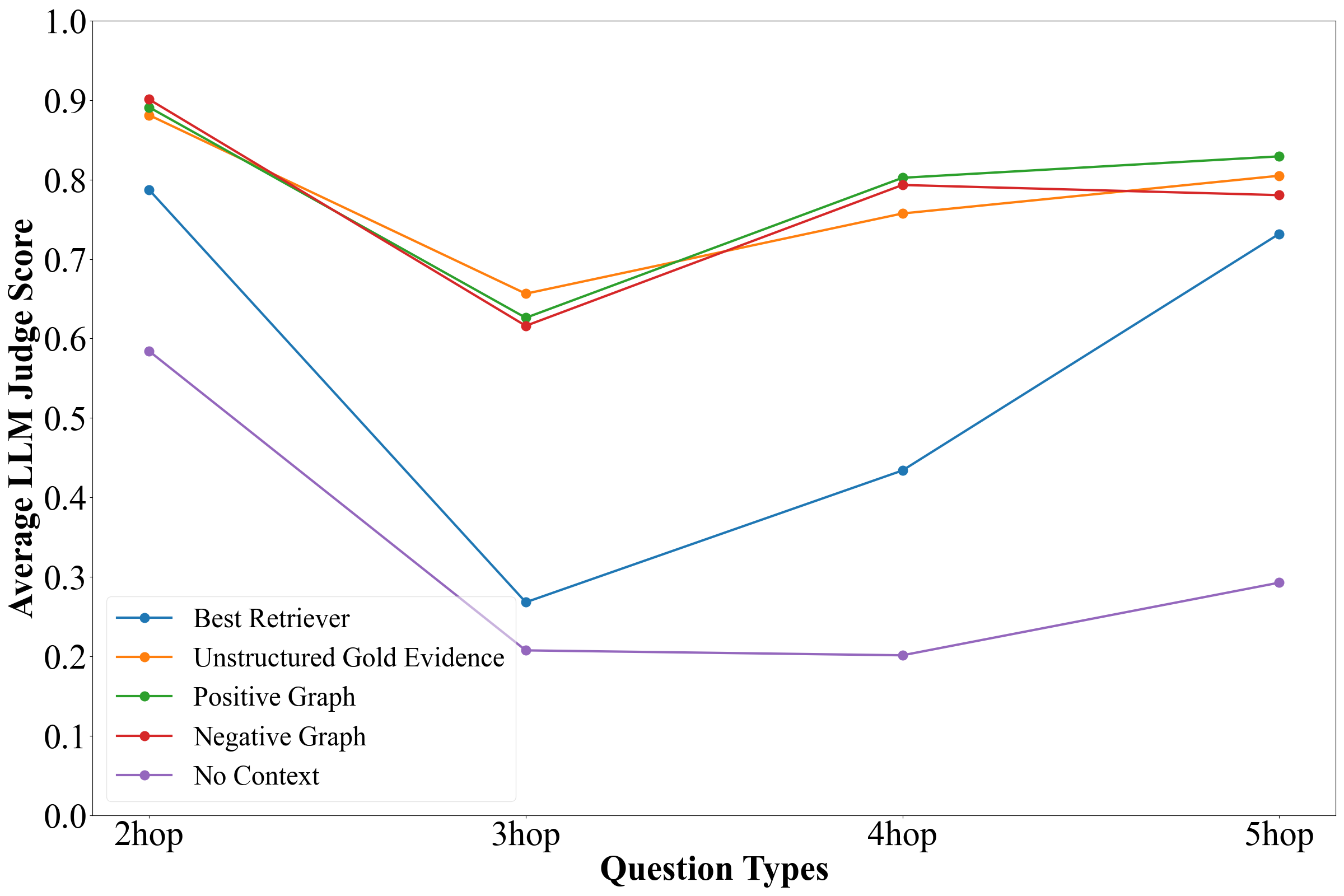}
        \caption{GPT-3.5 as LLM Judge}
        \label{fig:3.5_hop_type}
    \end{subfigure}
    \hfill
    \begin{subfigure}{0.32\textwidth}
        \centering
        \includegraphics[width=\linewidth]{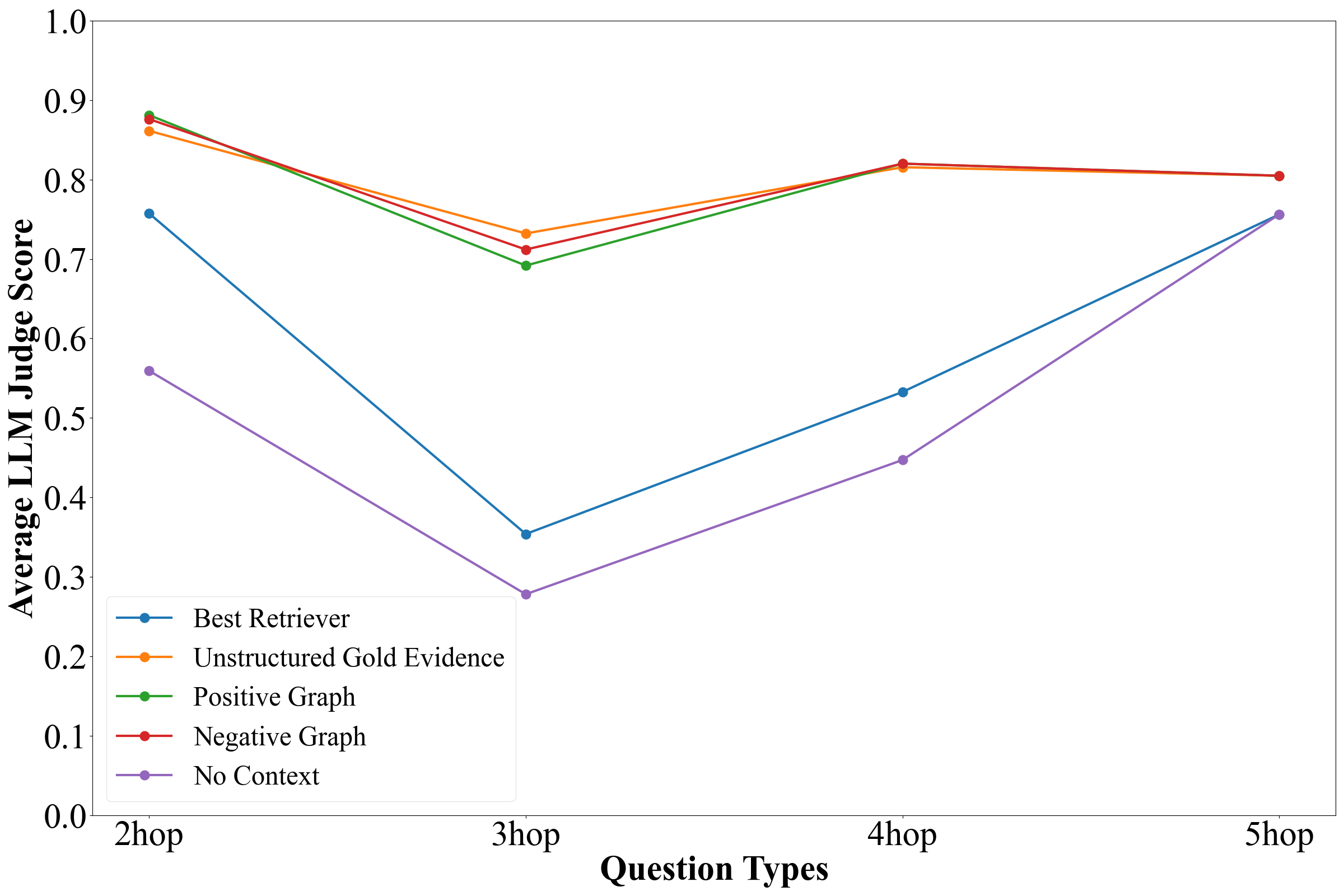}
        \caption{GPT-4o-mini as LLM Judge}
        \label{fig:4o_hop_type}
    \end{subfigure}
    \hfill
    \begin{subfigure}{0.32\textwidth}
        \centering
        \includegraphics[width=\linewidth]{images/llm_judge_scores_GPT-3.5.png}
        \caption{Llama3 as LLM Judge}
        \label{fig:llama_hop_type}
    \end{subfigure}
    \caption{LLM Judge Scores by Hop Type for Different LLMs}
    \label{fig:hop_types}
\end{figure*}

\subsection{LLM Performance Benchmark}

For this final benchmark, we explore four different retrieval configurations, best retriever, unstructured golden evidence, positive reasoning graphs, and negative reasoning graphs. First, we use our best retriever (BM25) and input the questions from the question set along with the top 20 retrieved edges. Second, we input the questions into the LLMs with unstructured gold evidence. Third, we provide the questions with structured gold evidence by using DOT language to replicate our positive reasoning graphs. Lastly, we prompt the LLMs with the questions and our negative reasoning graphs. For the performance of all three LLMs across these prompt types, refer to \autoref{fig:question_types} and \autoref{fig:hop_types}.

\subsubsection{RAG with Best Retriever}

For this experiment, we provide LLMs with the top 20 evidence sentences retrieved using BM25, which is identified as the best-performing retriever compared with TF-IDF and DPR. As shown in \autoref{tab:prompt0} in the Appendix, GPT-3.5 generally outperforms both Llama3 and GPT-4o-mini across most question types, particularly in simpler question types like bridge\_2\_1 and comparison\_2\_1, where GPT-3.5 achieves an F1 score of 0.70 and 0.78, respectively. However, all models struggle with more complex question types, such as compositional\_3\_2 and bridge\_4\_1, where the metrics were significantly lower. This experiment demonstrates that even with retrieval assistance, LLM performance drops as the question complexity increases. \autoref{fig:bm25_3.5}, \autoref{fig:bm25_4o-mini}, and \autoref{fig:bm25_llama3} show the results for each model on each question type with these added retrieval results.

\subsubsection{\hspace{-2ex} RAG with Unstructured Golden Evidence}

We then provide LLMs with unstructured golden evidence to investigate if removing irrelevant evidence improves performance. As expected, results improve significantly for all models across almost all question types. As seen in \autoref{tab:prompt1}, GPT-3.5 continues to lead with F1 scores as high as 1.00 for comparison\_4\_1. Even for more complex question types like bridge\_comparison\_5\_1, it achieves a strong F1 score of 0.74, indicating that access to relevant evidence boosts LLM performance. This improvement can be attributed to the reduction of noise from irrelevant evidence in the prompt. \autoref{fig:3.5_no_structure}, \autoref{fig:4o-mini_no_structure}, and \autoref{fig:llama3_no_structure} show the performance of each LLM for this set of experiments.

\subsubsection{RAG with Positive Reasoning Graph}

Building on the unstructured evidence prompting, we now prompt the LLMs with our positive reasoning graphs, mapping out the logical steps that guide us from question to answer. Similarly, we see an improvement in LLM performance, suggesting that reasoning graphs help models establish a clearer path from evidence to answer, improving both exact match and F1 scores across most question types.. \autoref{fig:3.5_pos}, \autoref{fig:4o-mini_pos} and \autoref{fig:llama3_pos} show the results for each LLM, while \autoref{tab:prompt2} contains the raw metric scores.

\subsubsection{RAG with Negative Reasoning Graph}

Lastly, we also evaluate the LLM's performance when prompted with a negative reasoning graph alongside the question. \autoref{fig:3.5_neg}, \autoref{fig:4o-mini_neg}, and \autoref{fig:llama3_neg} present the results for each LLM, while \autoref{tab:prompt3} provides the raw metric scores. In some cases, such as GPT-3.5 on the compositional\_4\_2 question type, we observe performance drops, along with other specific question types. This suggests that introducing negative reasoning graphs can sometimes lead to confusion or decreased performance, highlighting the sensitivity of LLMs to misleading or incorrect reasoning paths. This finding underscores the importance of ensuring that the reasoning graphs provided to LLMs are both accurate and relevant to avoid detrimental effects on their performance.

\section{Related Work}

In developing GRS-QA, we leverage insights from three influential multi-hop question-answering datasets: HotpotQA, MuSiQue, and 2WikiMultiHopQA. HotpotQA \cite{yang2018hotpotqa} introduced multi-hop reasoning, requiring models to draw from multiple documents to answer complex questions, with "supporting facts" marking essential context. However, models often bypass true multi-hop reasoning by exploiting shortcuts. MuSiQue \cite{trivedi2022musique} addressed this by designing questions that enforce "connected reasoning," ensuring models engage with all intermediate steps. 2WikiMultiHopQA \cite{ho2020constructing} further advanced multi-hop tasks by integrating structured and unstructured data from Wikipedia and Wikidata, emphasizing reasoning paths using evidence triples across four question types: inference, comparison, compositional, and bridge comparison. Building on these, our dataset challenges models with complex reasoning tasks, incorporating graph structures to prevent reasoning shortcuts. However, GRS-QA is proposing something that the above datasets lack: a structural representation of the reasoning path that is needed to arrive at an answer. This structural representation, although presented were not implemented in the datasets as we have done for GRS-QA. 

Complementary advancements in retrieval and reasoning also shaped GRS-QA. SURE  generates summaries from retrieved passages and verifies answers via zero-shot prompting \cite{kim2024sure}. \cite{karpukhin2020dense} uses dense retrieval with a dual-encoder for question and passage embeddings, while ORQAretrieves evidence using question-answer pairs for supervision \cite{lee2019latent}. \cite{izacard2021leveraging} combines generative models with passage retrieval. In KGQA, \cite{wu2023retrieverewriteanswer} transforms KG knowledge into text using a Rewriter module, and MRPQA enhances answer prediction with minimal labeled data \cite{wang-etal-2022-new}. \cite{chen-etal-2024-shot} offers a data synthesis framework with minimal human annotation, and PATHFID improves benchmarks by modeling reasoning paths \cite{yavuz2022modeling}. BeamQA uses beam search in the embedding space to enhance KGQA \cite{10.1145/3539618.3591698}. Recent LLM innovations include the LLM-Modulo framework for better critique reformulation in travel planning \cite{gundawar2024robust}, LLM-Planner for sample efficiency in visual environments \cite{Song_2023_ICCV}, and GLoRe, which boosts reasoning in LLaMA-2 with stepwise reward models \cite{havrilla2024glore}. The LLM Dynamic Planner excels in dynamic tasks \cite{dagan2023dynamic}, while LLM-Assist advances autonomous driving planning \cite{sharan2023llmassist}. Although these methods are improving with each release, GRS-QA poses an additional field that can be used to test models for their robustness when it comes to dealing with graph structures.

\section{Conclusion}
In this paper, we introduce \underline{G}raph \underline{R}easoning-\underline{S}tructured \underline{Q}uestion \underline{A}nswering Dataset (GRS-QA), that captures the different types of reasoning pathways necessary for multi-hop questions. Building on existing multi-hop QA datasets, such as HotpotQA, MuSiQue, and 2WikiMultiHopQA, we map out the logical steps required to reach each answer and construct reasoning graphs for every question. This structured representation provides a new lens for analyzing how LLMs handle complex multi-hop reasoning and offers a fine-grained evaluation of the reasoning steps.

Through our analysis of the reasoning structures in GRS-QA, we found that while current QA models perform well on many general tasks, they often struggle with questions that exhibit extremely complex reasoning structures, especially those requiring multi-step inference. This highlights the need for advancements in models’ ability to process and reason through intricate logical pathways.

\section{Limitations}

One key limitation of GRS-QA is the imbalanced distribution of graph types, as illustrated in Table\ref{table:reasoning-graphs}. Certain reasoning structures, such as bridge\_2\_1 (60.54\%) and comparison\_2\_1 (24.88\%), are heavily more, while more complex types like bridge\_comparison\_5\_1 and compositional\_4\_2 are significantly less. This imbalance may bias models toward performing better on more frequent graph types while struggling to generalize to less common but more complex reasoning patterns. 

In addition, the dataset spans multiple domains and does not adhere to a fixed subject area. While this diversity helps increase the generalization capabilities of models trained on the dataset, it introduces challenges in domain-specific reasoning. Questions may vary significantly in their topics, from historical facts to scientific inquiries, without clear demarcation.

Lastly, while the inclusion of multi-hop reasoning graphs is a strength, it also presents a limitation. The complex nature of these graphs requires models to engage in intricate, multi-step reasoning, which is not yet fully optimized in current LLMs.

\subsection{Future Work}
As mentioned above, one of the limitations of this dataset is the imbalance of data types based on the complexity of the reasoning graph. In order to combat this, synthetic data could be generated to duplicate the complex structures that are being underrepresented in the current data. Given that GRS-QA spans multiple domains, a promising direction for future work would be to segment the dataset by domain (e.g., historical, scientific, general knowledge) and develop domain-adapted models or assist in creating models that are more robust to varying domains. This could help improve the performance of QA models in domain-specific tasks, allowing for more accurate reasoning in specialized contexts. Furthermore, integrating domain-specific knowledge bases could enhance multi-hop reasoning, enabling models to better handle the diversity of content.

Furthermore, expanding the variety and quantity of these negative graphs could offer deeper insights into how different graph structures influence or impede the performance of LLMs in completing tasks. By increasing the diversity of negative graph structures, we can better understand how these models navigate and respond to complex reasoning challenges, potentially uncovering patterns that enhance or hinder their reasoning capabilities.

Finally, we plan to expand our benchmarking efforts by testing a wider variety of model architectures on the dataset. Comparing the performance of different approaches, such as graph neural networks (GNNs) or retrieval-augmented models. It could provide insights into which types of architectures are best suited to handle graph structure

\bibliographystyle{acl_natbib}
\bibliography{acl_latex}

\appendix
\section{Dataset Processing}

\subsection{HotPotQA Processing}
For each data item, nodes and edges are constructed, with nodes generated from positive paragraphs and characterized by attributes such as evidence and type. In cases where the data item is classified as a "bridge" type, specific handling is applied to determine the presence of a "bridge" in the positive paragraphs, influencing edge connections. This results in a primary graph structure containing nodes and edges. The process also generates 5 sets of negative examples, where each set consists of two graphs: one with an additional node and another with altered edge connections, creating challenging negative samples. Additionally, another five sets of negative examples are created by randomly selecting nodes from negative paragraphs and constructing corresponding edges. The final output comprises a list of processed data items, including the original question, concatenated positive evidence, correct answers, the primary positive graph, and sets of negative graphs.

\subsection{MuSiQue Processing}
For each data item, nodes and edges are constructed, with nodes generated from the answers of the question decomposition and edges created by using attributes such as the type. The process also generates five sets of negative examples, where each set consists of two graphs: one with an additional node and another with altered edge connections, creating challenging negative samples. Additionally, another five sets of negative examples are created by randomly selecting nodes from negative paragraphs and constructing corresponding edges. The final output comprises a list of processed data items, including the original question, concatenated positive evidence, correct answers, the primary positive graph, and sets of hard and easy negative graphs.

\subsection{2WikiMultiHopQA Processing}
To construct 2WikiMultiHopQA reasoning graphs, we extracted four primary graph types—Inference, Comparison, Bridge Comparison, and Compositional—from the raw dataset, varying node counts to generate additional structures. Gold graphs were built using supporting facts and evidences, with nodes representing supporting facts and edges created based on relations between entities shown in the evidences. Five sets of negative graphs were created by using nodes from the context to add an extra node. Another five sets of negative graphs were created by randomly selecting and constructing the corresponding edges. The final output comprises a list of processed data items, including the original question, concatenated positive evidence, correct answers, the primary positive graph, and sets of negative graphs.

Due to entity string inconsistencies in the data for 2WikiMultiHopQA train set, there were 4507 instances of the data that were discarded since they were unable to be used to create positive graphs.  

\subsection{Licensing}

Our use of the following datasets is consistent with their licenses, specifically:

\begin{itemize}
    \item 2WikiMultiHopQA: Apache-2.0 License \href{https://github.com/Alab-NII/2wikimultihop?tab=Apache-2.0-1-ov-file}{https://github.com/Alab-NII/2wikimultihop?tab=Apache-2.0-1-ov-file}
    \item HotpotQA: CC BY-SA 4.0 License \href{https://hotpotqa.github.io/}{https://hotpotqa.github.io/}
    \item MuSiQue: CC BY 4.0 License \href{https://github.com/stonybrooknlp/musique}{https://github.com/stonybrooknlp/musique}
\end{itemize}
Our dataset will be released under the Creative Commons Attribution 4.0 International License (CC BY 4.0).

\section{Statistical Analysis -- Breakdown of each dataset}

\begin{table}[htbp]
\centering
\begin{tabular}{lccc}
\toprule
\textbf{Question Type} & \textbf{Train} & \textbf{Val} & \textbf{Test} \\
\midrule
Bridge\_2\_1 & 58384 & 7298 & 7298 \\
Comparison\_2\_1 & 13964 & 1745 & 1747 \\
\midrule
\textbf{total} & \textbf{72348} & \textbf{9043} & \textbf{9045}\\
\bottomrule
\end{tabular}
\caption{Breakdown of Question Types and Unique Question Count for HotpotQA}
\end{table}

\begin{table}[htbp]
\centering
\begin{tabular}{lccc}
\toprule
\textbf{Question Type} & \textbf{Train} & \textbf{Val} & \textbf{Test} \\
\midrule
Bridge\_2\_1 & 61209 & 7651 & 7652 \\
Comparison\_2\_1 & 41324 & 5165 & 5167 \\
Comparison\_3\_1 & 234 & 29 & 30 \\
Comparison\_4\_1 & 10 & 1 & 2 \\
Comparison\_5\_1 & - & - & 1 \\
Compositional\_3\_2 & 3 & - & 1 \\
Bridge\_Comparison\_4\_1 & 27266 & 3408 & 3409 \\
Bridge\_Comparison\_5\_1 & 308 & 38 & 29 \\
\midrule
\textbf{total} & \textbf{130354} & \textbf{16292} & \textbf{16301}\\
\bottomrule
\end{tabular}
\caption{Breakdown of Question Types and Unique Question Count for 2WikiMultiHopQA}
\end{table}

\begin{table}[htbp]
\centering
\begin{tabular}{lccc}
\hline
\textbf{Question Type} & \textbf{Train} & \textbf{Val} & \textbf{Test} \\
\hline
Bridge\_2\_1  & 11478 & 1434 & 1436 \\
Bridge\_3\_1 & 2987 & 373 & 374 \\
Compositional\_3\_2 & 519 & 64 & 66 \\
Bridge\_4\_1 & 516 & 64 & 65 \\
Compositional\_4\_2 & 101 & 12 & 14 \\
Compositional\_4\_3 & 319 & 39 & 41 \\
\hline
\textbf{total} & \textbf{15920} & \textbf{1986} & \textbf{1996}\\
\hline
\end{tabular}
\caption{Breakdown of Question Types and Unique Question Count for MuSiQue}
\end{table}

\section{Experiments Tables and Graphs}\label{sec:experiments_appendix}
\begin{table}[htbp]
\centering
\begin{tabular}{lccc}
\hline
\textbf{Method} & \textbf{Recall} & \textbf{F1} & \textbf{Precision} \\ \hline
BM25   & \textbf{0.4921} & \textbf{0.1182} & \textbf{0.0680} \\ 
TF-IDF & 0.1619 & 0.0447 & 0.0261 \\ 
DPR    & 0.1037 & 0.0285 & 0.0166 \\ \hline
\end{tabular}
\caption{Average Retrieval Metrics for BM25, TF-IDF, and DPR}
\label{tab:retrieval_metrics}
\end{table}

\begin{table*}[htbp]
\centering
\makebox[\textwidth][c]{
\resizebox{1.2\textwidth}{!}{
\begin{tabular}{lccccccccc}
\hline
Graph Type & EM (Llama3-8B) & F1 (Llama3-8B) & LLM (Llama3-8B) & EM (GPT-4o-mini) & F1 (GPT-4o-mini) & LLM (GPT-4o-mini) & EM (GPT-3.5) & F1 (GPT-3.5) & LLM (GPT-3.5) \\ 
\hline
bridge\_2\_1 & 0.49 & 0.60 & 0.66 & 0.50 & 0.67 & 0.73 & 0.58 & 0.70 & 0.72 \\ 
comparison\_2\_1 & 0.43 & 0.53 & 0.75 & 0.46 & 0.58 & 0.78 & 0.70 & 0.78 & 0.85 \\ 
bridge\_3\_1 & 0.04 & 0.07 & 0.09 & 0.10 & 0.23 & 0.28 & 0.11 & 0.20 & 0.20 \\ 
comparison\_3\_1 & 0.47 & 0.50 & 0.63 & 0.07 & 0.25 & 0.70 & 0.57 & 0.58 & 0.63 \\ 
compositional\_3\_2 & 0.04 & 0.06 & 0.10 & 0.07 & 0.15 & 0.31 & 0.06 & 0.10 & 0.21 \\ 
bridge\_4\_1 & 0.03 & 0.06 & 0.15 & 0.06 & 0.19 & 0.34 & 0.08 & 0.17 & 0.18 \\ 
comparison\_4\_1 & 0.50 & 0.57 & 1.00 & 0.00 & 0.17 & 0.50 & 0.50 & 0.50 & 0.50 \\ 
compositional\_4\_2 & 0.00 & 0.04 & 0.07 & 0.07 & 0.09 & 0.14 & 0.07 & 0.07 & 0.07 \\ 
compositional\_4\_3 & 0.15 & 0.16 & 0.22 & 0.22 & 0.28 & 0.46 & 0.22 & 0.24 & 0.37 \\ 
bridge\_comparison\_4\_1 & 0.30 & 0.40 & 0.66 & 0.20 & 0.37 & 0.74 & 0.63 & 0.66 & 0.67 \\ 
comparison\_5\_1 & 0.00 & 0.08 & 0.00 & 0.00 & 0.00 & 0.00 & 0.00 & 0.00 & 0.00 \\ 
bridge\_comparison\_5\_1 & 0.50 & 0.52 & 0.72 & 0.00 & 0.26 & 0.78 & 0.70 & 0.71 & 0.75 \\ 
\hline
\end{tabular}
}
}
\caption{Metrics for Prompt Type 0: Question w/ Top-20 Retrieved Evidence Using BM25}
\label{tab:prompt0}
\end{table*}

\begin{table*}[htbp]
\centering
\makebox[\textwidth][c]{
\resizebox{1.2\textwidth}{!}{
\begin{tabular}{lccccccccc}
\hline
Graph Type & EM (Llama3-8B) & F1 (Llama3-8B) & LLM (Llama3-8B) & EM (GPT-4o-mini) & F1 (GPT-4o-mini) & LLM (GPT-4o-mini) & EM (GPT-3.5) & F1 (GPT-3.5) & LLM (GPT-3.5) \\ 
\hline
bridge\_2\_1 & 0.59 & 0.77 & 0.86 & 0.69 & 0.85 & 0.90 & 0.65 & 0.80 & 0.90 \\ 
comparison\_2\_1 & 0.32 & 0.48 & 0.82 & 0.58 & 0.68 & 0.82 & 0.54 & 0.67 & 0.86 \\ 
bridge\_3\_1 & 0.31 & 0.44 & 0.62 & 0.45 & 0.63 & 0.75 & 0.36 & 0.50 & 0.68 \\ 
comparison\_3\_1 & 0.43 & 0.49 & 0.70 & 0.57 & 0.63 & 0.73 & 0.60 & 0.64 & 0.67 \\ 
compositional\_3\_2 & 0.21 & 0.31 & 0.54 & 0.39 & 0.52 & 0.70 & 0.30 & 0.40 & 0.61 \\ 
bridge\_4\_1 & 0.18 & 0.35 & 0.63 & 0.42 & 0.62 & 0.71 & 0.32 & 0.54 & 0.68 \\ 
comparison\_4\_1 & 0.50 & 0.61 & 1.00 & 1.00 & 1.00 & 1.00 & 1.00 & 1.00 & 1.00 \\ 
compositional\_4\_2 & 0.21 & 0.35 & 0.64 & 0.64 & 0.71 & 0.86 & 0.50 & 0.58 & 0.64 \\ 
compositional\_4\_3 & 0.32 & 0.38 & 0.61 & 0.39 & 0.44 & 0.68 & 0.37 & 0.41 & 0.66 \\ 
bridge\_comparison\_4\_1 & 0.09 & 0.33 & 0.89 & 0.66 & 0.76 & 0.93 & 0.77 & 0.80 & 0.86 \\ 
comparison\_5\_1 & 0.00 & 0.06 & 0.00 & 1.00 & 1.00 & 1.00 & 0.00 & 0.40 & 1.00 \\ 
bridge\_comparison\_5\_1 & 0.05 & 0.18 & 0.82 & 0.38 & 0.53 & 0.80 & 0.70 & 0.74 & 0.80 \\ 
\hline
\end{tabular}
}}
\caption{Metrics for Prompt Type 1: Question w/ Unstructured Ground Truth Evidence}
\label{tab:prompt1}
\end{table*}

\begin{table*}[htbp]
\centering
\makebox[\textwidth][c]{
\resizebox{1.2\textwidth}{!}{
\begin{tabular}{lccccccccc}
\hline
Graph Type & EM (Llama3-8B) & F1 (Llama3-8B) & LLM (Llama3-8B) & EM (GPT-4o-mini) & F1 (GPT-4o-mini) & LLM (GPT-4o-mini) & EM (GPT-3.5) & F1 (GPT-3.5) & LLM (GPT-3.5) \\ 
\hline
bridge\_2\_1 & 0.66 & 0.83 & 0.88 & 0.69 & 0.85 & 0.90 & 0.68 & 0.82 & 0.88 \\ 
comparison\_2\_1 & 0.34 & 0.51 & 0.84 & 0.50 & 0.64 & 0.86 & 0.52 & 0.68 & 0.90 \\ 
bridge\_3\_1 & 0.38 & 0.52 & 0.66 & 0.45 & 0.64 & 0.71 & 0.38 & 0.52 & 0.66 \\ 
comparison\_3\_1 & 0.43 & 0.49 & 0.67 & 0.37 & 0.47 & 0.70 & 0.60 & 0.63 & 0.67 \\ 
compositional\_3\_2 & 0.22 & 0.32 & 0.61 & 0.39 & 0.51 & 0.66 & 0.27 & 0.37 & 0.55 \\ 
bridge\_4\_1 & 0.20 & 0.40 & 0.68 & 0.42 & 0.63 & 0.68 & 0.38 & 0.62 & 0.72 \\ 
comparison\_4\_1 & 1.00 & 1.00 & 1.00 & 0.50 & 0.70 & 1.00 & 1.00 & 1.00 & 1.00 \\ 
compositional\_4\_2 & 0.29 & 0.41 & 0.57 & 0.71 & 0.75 & 0.79 & 0.64 & 0.68 & 0.79 \\ 
compositional\_4\_3 & 0.32 & 0.37 & 0.63 & 0.49 & 0.55 & 0.78 & 0.34 & 0.44 & 0.71 \\ 
bridge\_comparison\_4\_1 & 0.07 & 0.33 & 0.90 & 0.41 & 0.58 & 0.93 & 0.70 & 0.78 & 0.89 \\ 
comparison\_5\_1 & 0.00 & 0.17 & 1.00 & 1.00 & 1.00 & 1.00 & 0.00 & 0.40 & 1.00 \\ 
bridge\_comparison\_5\_1 & 0.03 & 0.20 & 0.78 & 0.07 & 0.31 & 0.80 & 0.65 & 0.71 & 0.82 \\ 
\hline
\end{tabular}
}}
\caption{Metrics for Prompt Type 2: Question Positive Graph of Ground Truth Evidence}
\label{tab:prompt2}
\end{table*}

\begin{table*}[htbp]
\centering
\makebox[\textwidth][c]{
\resizebox{1.2\textwidth}{!}{
\begin{tabular}{lccccccccc}
\hline
Graph Type & EM (Llama3-8B) & F1 (Llama3-8B) & LLM (Llama3-8B) & EM (GPT-4o-mini) & F1 (GPT-4o-mini) & LLM (GPT-4o-mini) & EM (GPT-3.5) & F1 (GPT-3.5) & LLM (GPT-3.5) \\ 
\hline
bridge\_2\_1 & 0.61 & 0.80 & 0.87 & 0.69 & 0.85 & 0.89 & 0.68 & 0.83 & 0.91 \\ 
comparison\_2\_1 & 0.35 & 0.52 & 0.83 & 0.50 & 0.64 & 0.86 & 0.52 & 0.68 & 0.89 \\ 
bridge\_3\_1 & 0.40 & 0.54 & 0.66 & 0.46 & 0.63 & 0.73 & 0.41 & 0.54 & 0.65 \\ 
comparison\_3\_1 & 0.47 & 0.51 & 0.63 & 0.33 & 0.45 & 0.70 & 0.53 & 0.57 & 0.63 \\ 
compositional\_3\_2 & 0.27 & 0.36 & 0.63 & 0.43 & 0.54 & 0.69 & 0.24 & 0.36 & 0.55 \\ 
bridge\_4\_1 & 0.25 & 0.42 & 0.65 & 0.45 & 0.64 & 0.68 & 0.37 & 0.58 & 0.69 \\ 
comparison\_4\_1 & 1.00 & 1.00 & 1.00 & 0.50 & 0.70 & 1.00 & 1.00 & 1.00 & 1.00 \\ 
compositional\_4\_2 & 0.29 & 0.43 & 0.71 & 0.71 & 0.75 & 0.79 & 0.64 & 0.66 & 0.71 \\ 
compositional\_4\_3 & 0.29 & 0.37 & 0.63 & 0.49 & 0.54 & 0.78 & 0.34 & 0.43 & 0.71 \\ 
bridge\_comparison\_4\_1 & 0.08 & 0.32 & 0.91 & 0.39 & 0.56 & 0.93 & 0.68 & 0.76 & 0.90 \\ 
comparison\_5\_1 & 0.00 & 0.18 & 1.00 & 1.00 & 1.00 & 1.00 & 0.00 & 0.40 & 1.00 \\ 
bridge\_comparison\_5\_1 & 0.03 & 0.18 & 0.78 & 0.05 & 0.29 & 0.80 & 0.65 & 0.70 & 0.78 \\ 
\hline
\end{tabular}
}}
\caption{Metrics for Prompt Type 3: Question w/ Negative Graph of Ground Truth Evidence}
\label{tab:prompt3}
\end{table*}

\begin{table*}[htbp]
\centering
\makebox[\textwidth][c]{
\resizebox{1.2\textwidth}{!}{
\begin{tabular}{lccccccccc}
\hline
Graph Type & EM (Llama3-8B) & F1 (Llama3-8B) & LLM (Llama3-8B) & EM (GPT-4o-mini) & F1 (GPT-4o-mini) & LLM (GPT-4o-mini) & EM (GPT-3.5) & F1 (GPT-3.5) & LLM (GPT-3.5) \\ 
\hline
bridge\_2\_1 & 0.17 & 0.27 & 0.33 & 0.31 & 0.40 & 0.39 & 0.27 & 0.38 & 0.46 \\ 
comparison\_2\_1 & 0.25 & 0.38 & 0.64 & 0.49 & 0.57 & 0.73 & 0.52 & 0.62 & 0.71 \\ 
bridge\_3\_1 & 0.04 & 0.10 & 0.12 & 0.11 & 0.21 & 0.21 & 0.09 & 0.18 & 0.13 \\ 
comparison\_3\_1 & 0.40 & 0.47 & 0.63 & 0.17 & 0.32 & 0.63 & 0.47 & 0.52 & 0.63 \\ 
compositional\_3\_2 & 0.01 & 0.04 & 0.09 & 0.09 & 0.16 & 0.22 & 0.03 & 0.09 & 0.13 \\ 
bridge\_4\_1 & 0.00 & 0.05 & 0.12 & 0.02 & 0.11 & 0.20 & 0.02 & 0.07 & 0.11 \\ 
comparison\_4\_1 & 0.50 & 0.50 & 0.50 & 0.00 & 0.12 & 0.50 & 0.00 & 0.00 & 0.00 \\ 
compositional\_4\_2 & 0.00 & 0.06 & 0.14 & 0.00 & 0.14 & 0.14 & 0.07 & 0.07 & 0.07 \\ 
compositional\_4\_3 & 0.05 & 0.07 & 0.10 & 0.10 & 0.14 & 0.20 & 0.02 & 0.08 & 0.12 \\ 
bridge\_comparison\_4\_1 & 0.02 & 0.23 & 0.56 & 0.40 & 0.52 & 0.75 & 0.14 & 0.21 & 0.32 \\ 
comparison\_5\_1 & 0.00 & 0.22 & 1.00 & 1.00 & 1.00 & 1.00 & 1.00 & 1.00 & 1.00 \\ 
bridge\_comparison\_5\_1 & 0.00 & 0.17 & 0.68 & 0.23 & 0.40 & 0.75 & 0.15 & 0.18 & 0.28 \\ 
\hline
\end{tabular}
}}
\caption{Metrics for Prompt Type 4: Question w/ No Evidence}
\label{tab:prompt4}
\end{table*}

\begin{figure}[htbp]
    \centering
    \includegraphics[width=1\linewidth]{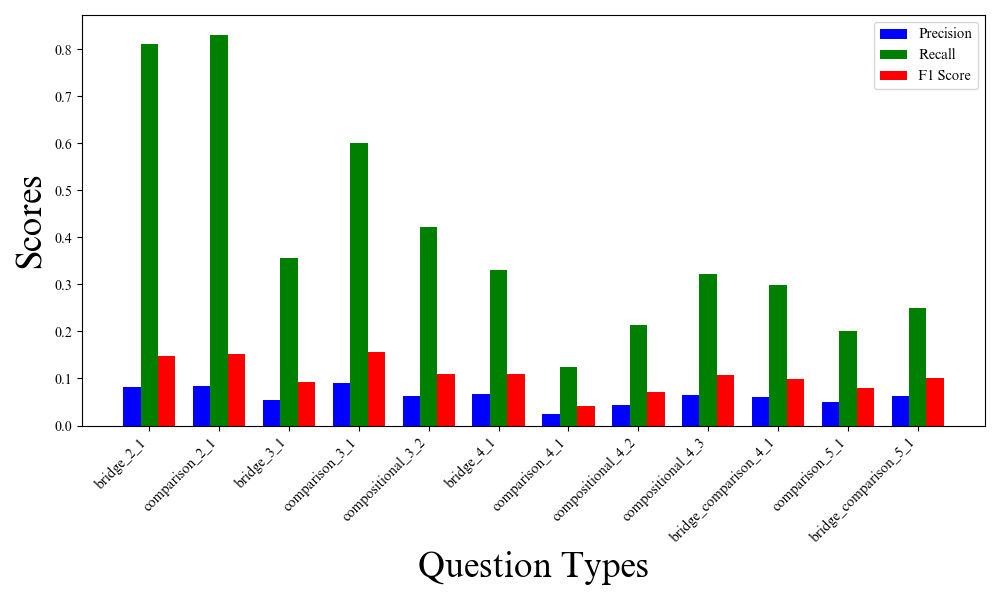}
    \caption{BM25 Retrieval Across Question Types}
    \label{fig:bm25_retrieval}
\end{figure}

\begin{figure}[htbp]
    \centering
    \includegraphics[width=1\linewidth]{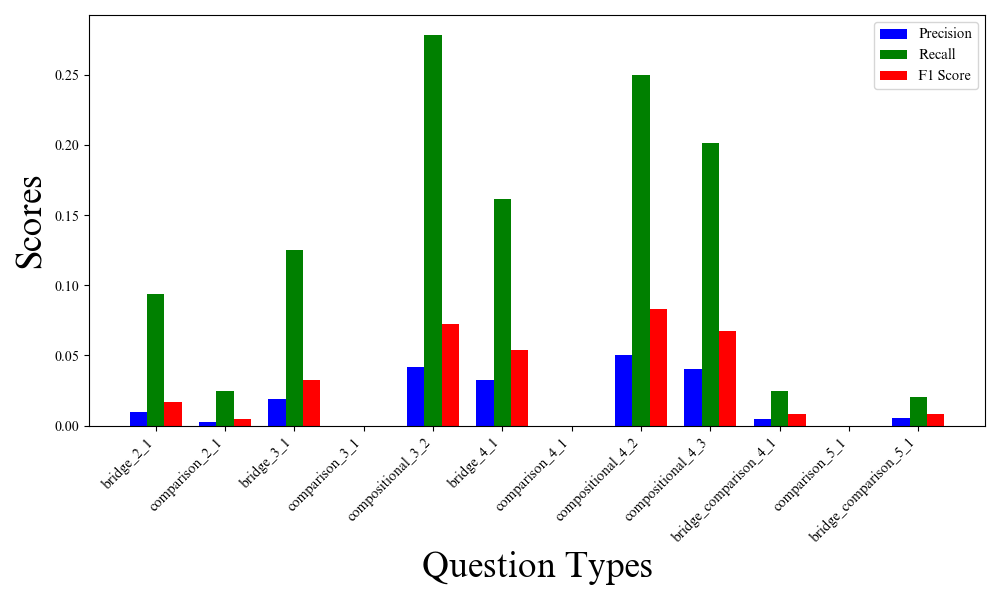}
    \caption{DPR Retrieval Across Question Types}
    \label{fig:dpr_retrieval}
\end{figure}

\begin{figure}[htbp]
    \centering
    \includegraphics[width=1\linewidth]{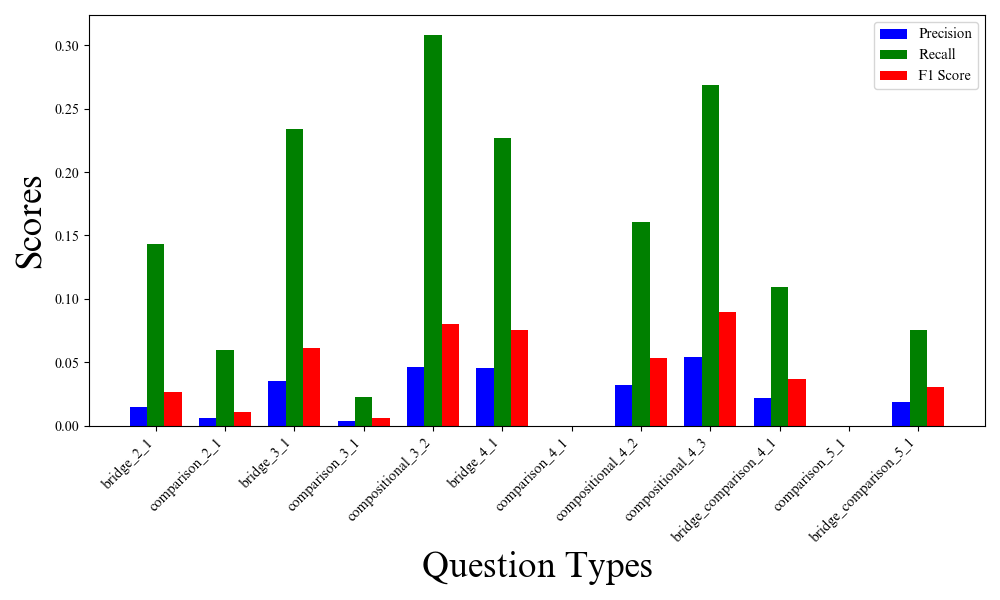}
    \caption{TFIDF Retrieval Across Question Types}
    \label{fig:tfidf_retrieval}
\end{figure}

\begin{figure}[htbp]
    \centering
    \includegraphics[width=1\linewidth]{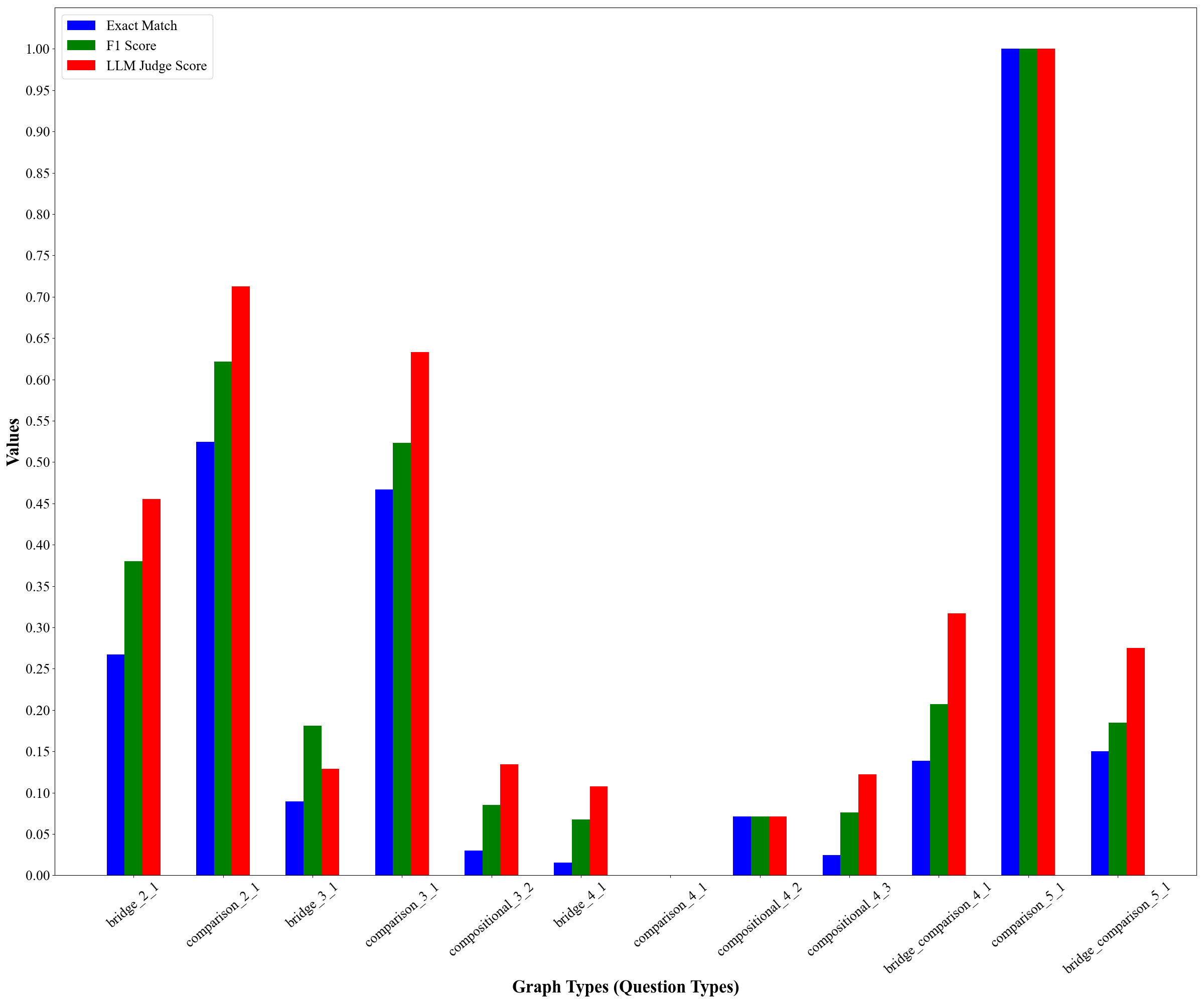}
    \caption{GPT-3.5 Metrics - No Context Provided}
    \label{fig:3.5_direct}
\end{figure}

\begin{figure}[htbp]
    \centering
    \includegraphics[width=1\linewidth]{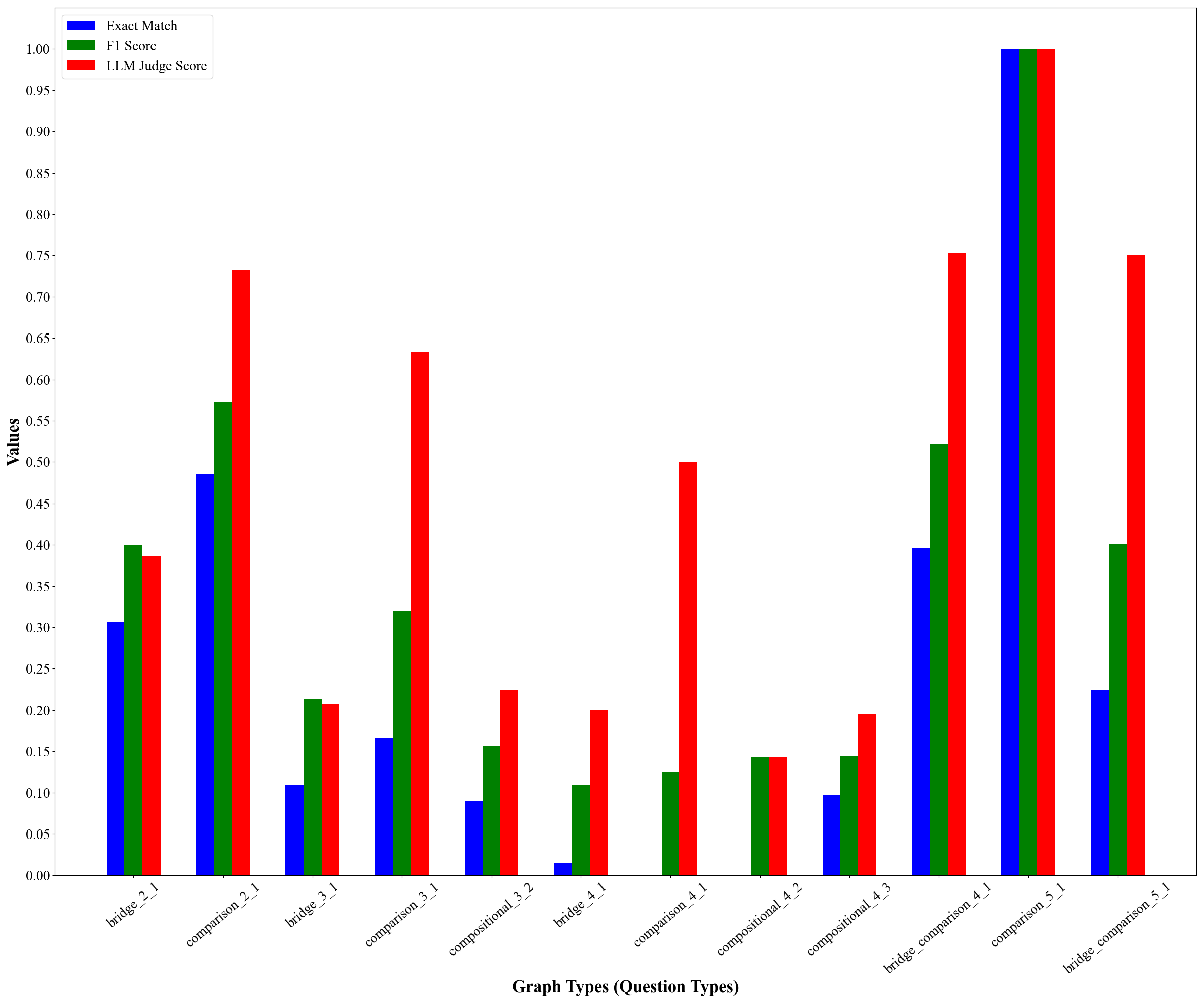}
    \caption{GPT4o-mini Metrics - No Context Provided}
    \label{fig:4o-mini_direct}
\end{figure}

\begin{figure}[htbp]
    \centering
    \includegraphics[width=1\linewidth]{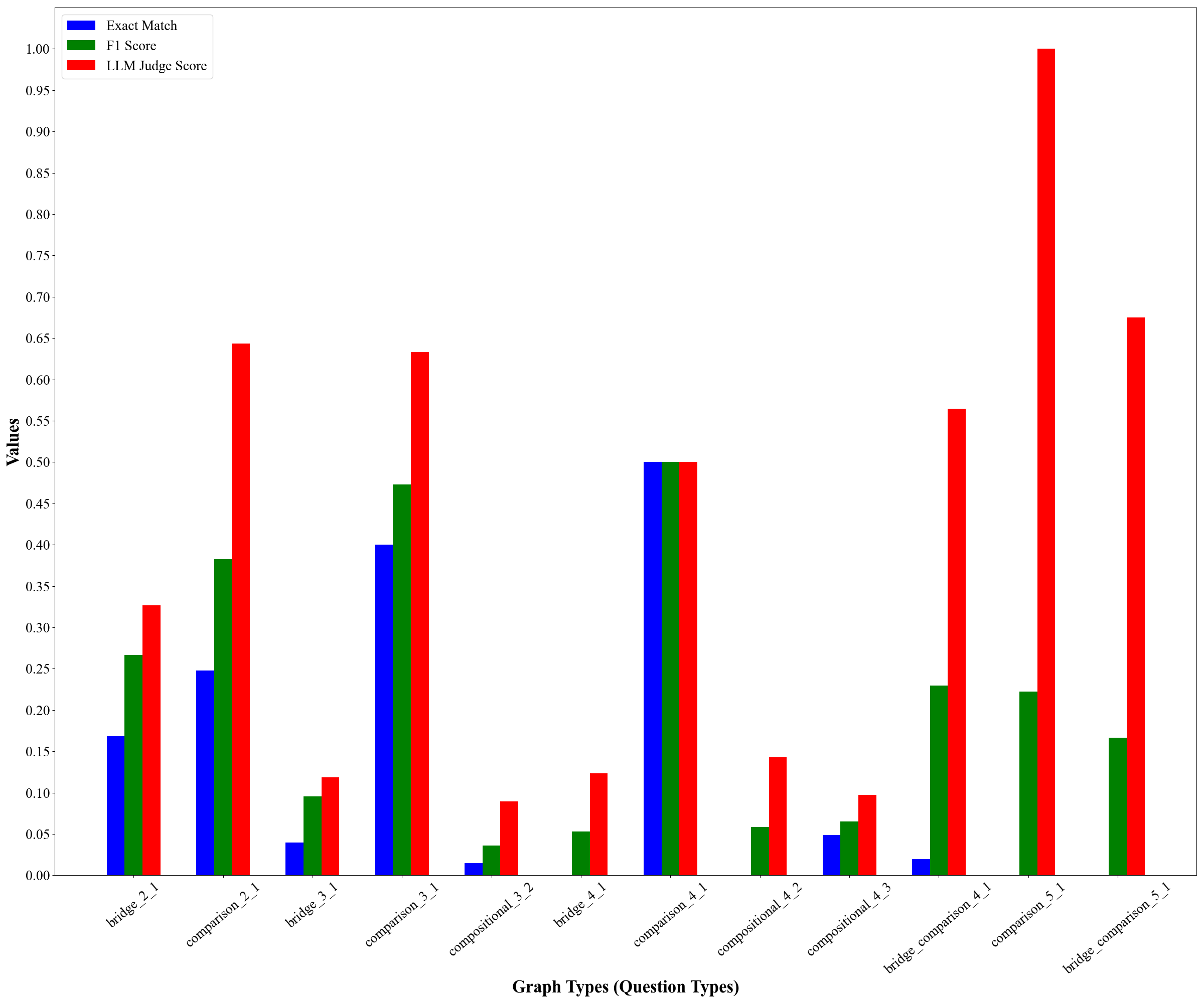}
    \caption{Llama3 Metrics - No Context Provided}
    \label{fig:llama3_direct}
\end{figure}

\begin{figure}[htbp]
    \centering
    \includegraphics[width=1\linewidth]{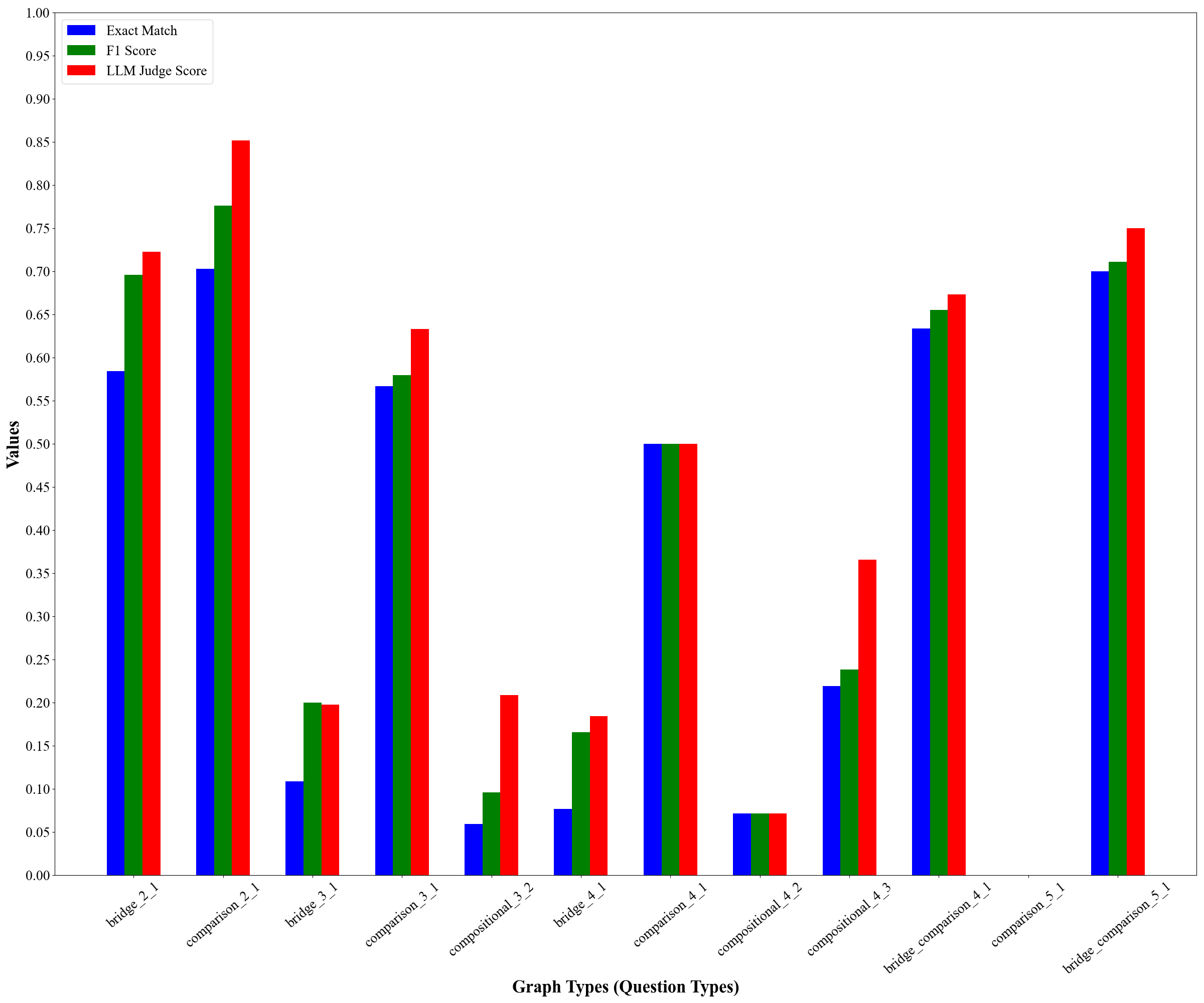}
    \caption{GPT-3.5 Metrics - Best Retriever}
    \label{fig:bm25_3.5}
\end{figure}

\begin{figure}[htbp]
    \centering
    \includegraphics[width=1\linewidth]{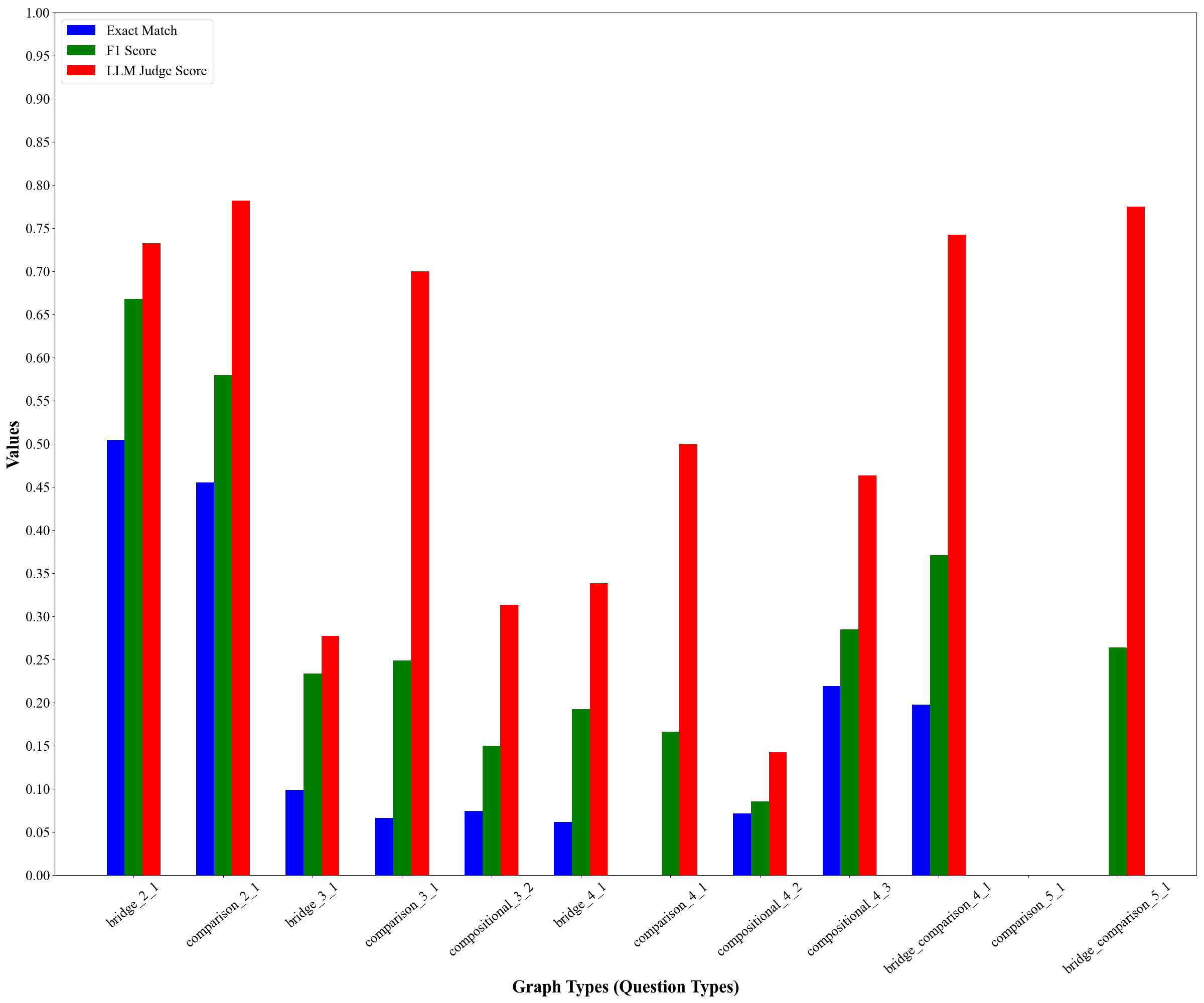}
    \caption{GPT4o-mini Metrics - Best Retriever}
    \label{fig:bm25_4o-mini}
\end{figure}

\begin{figure}[htbp]
    \centering
    \includegraphics[width=1\linewidth]{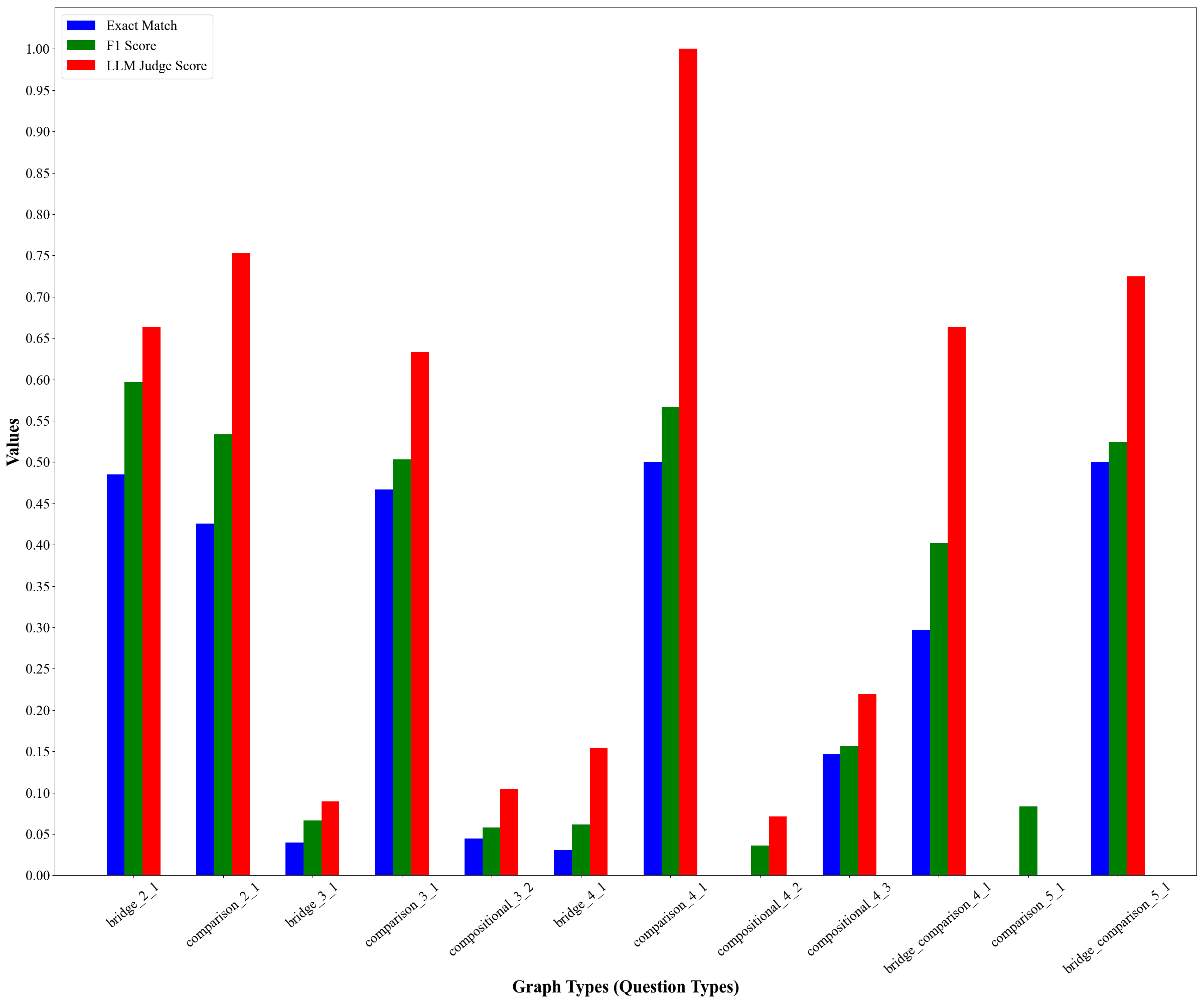}
    \caption{Llama3 Metrics - Best Retriever}
    \label{fig:bm25_llama3}
\end{figure}

\begin{figure}[htbp]
    \centering
    \includegraphics[width=1\linewidth]{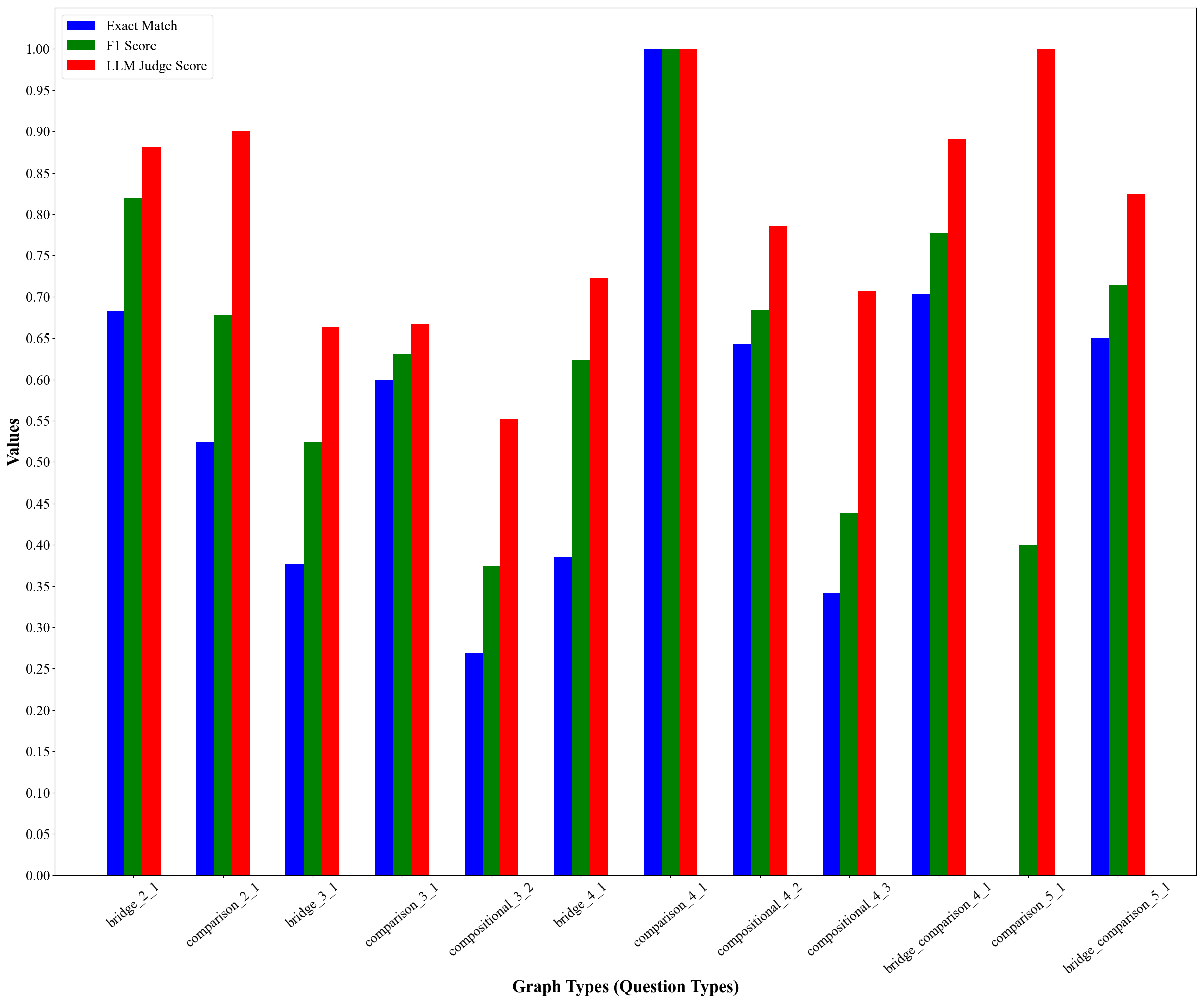}
    \caption{GPT-3.5 Metrics - Positive Graph of Ground Truth Evidence}
    \label{fig:3.5_pos}
\end{figure}

\begin{figure}[htbp]
    \centering
    \includegraphics[width=1\linewidth]{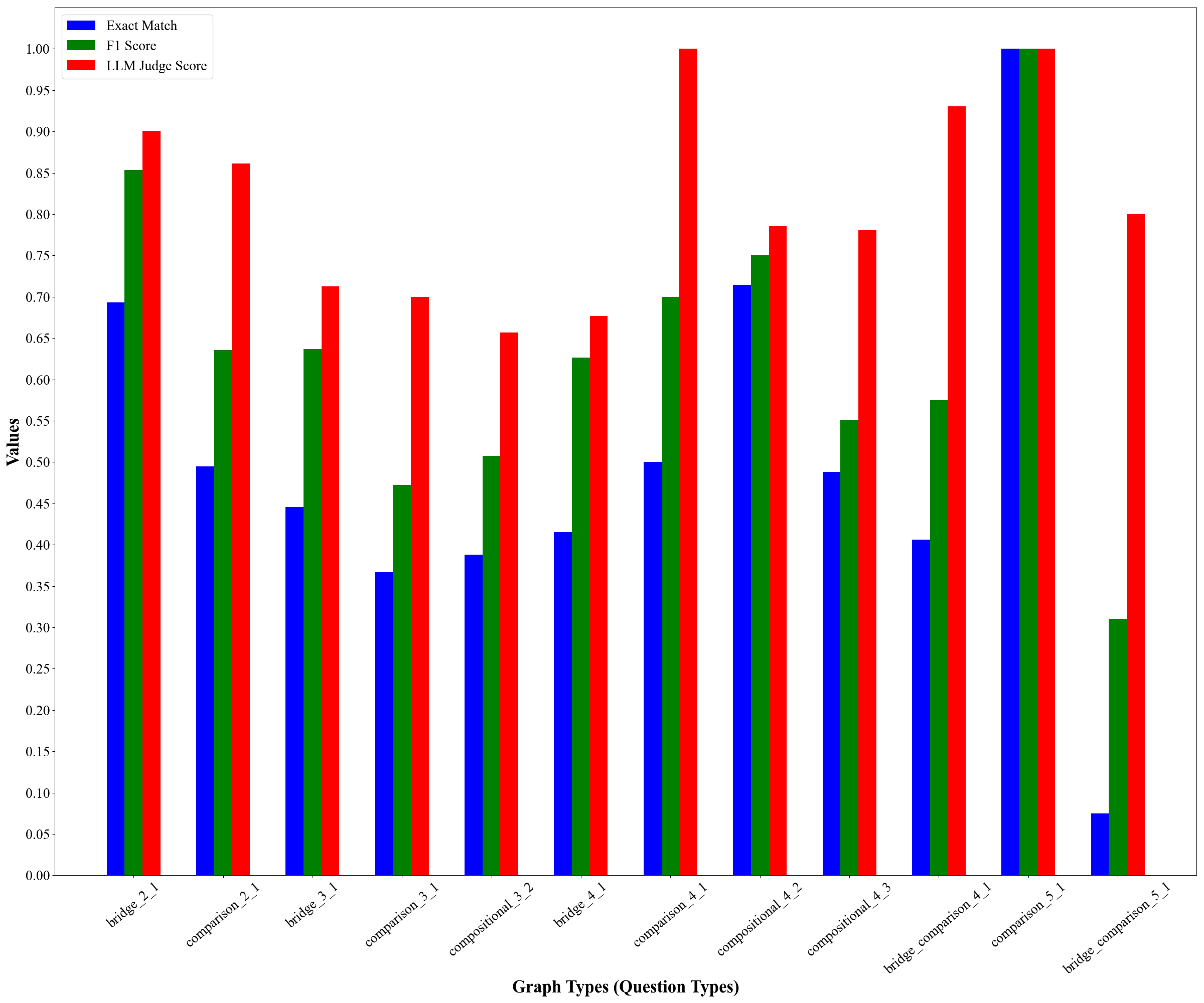}
    \caption{GPT4o-mini Metrics - Positive Graph of Ground Truth Evidence}
    \label{fig:4o-mini_pos}
\end{figure}

\begin{figure}[htbp]
    \centering
    \includegraphics[width=1\linewidth]{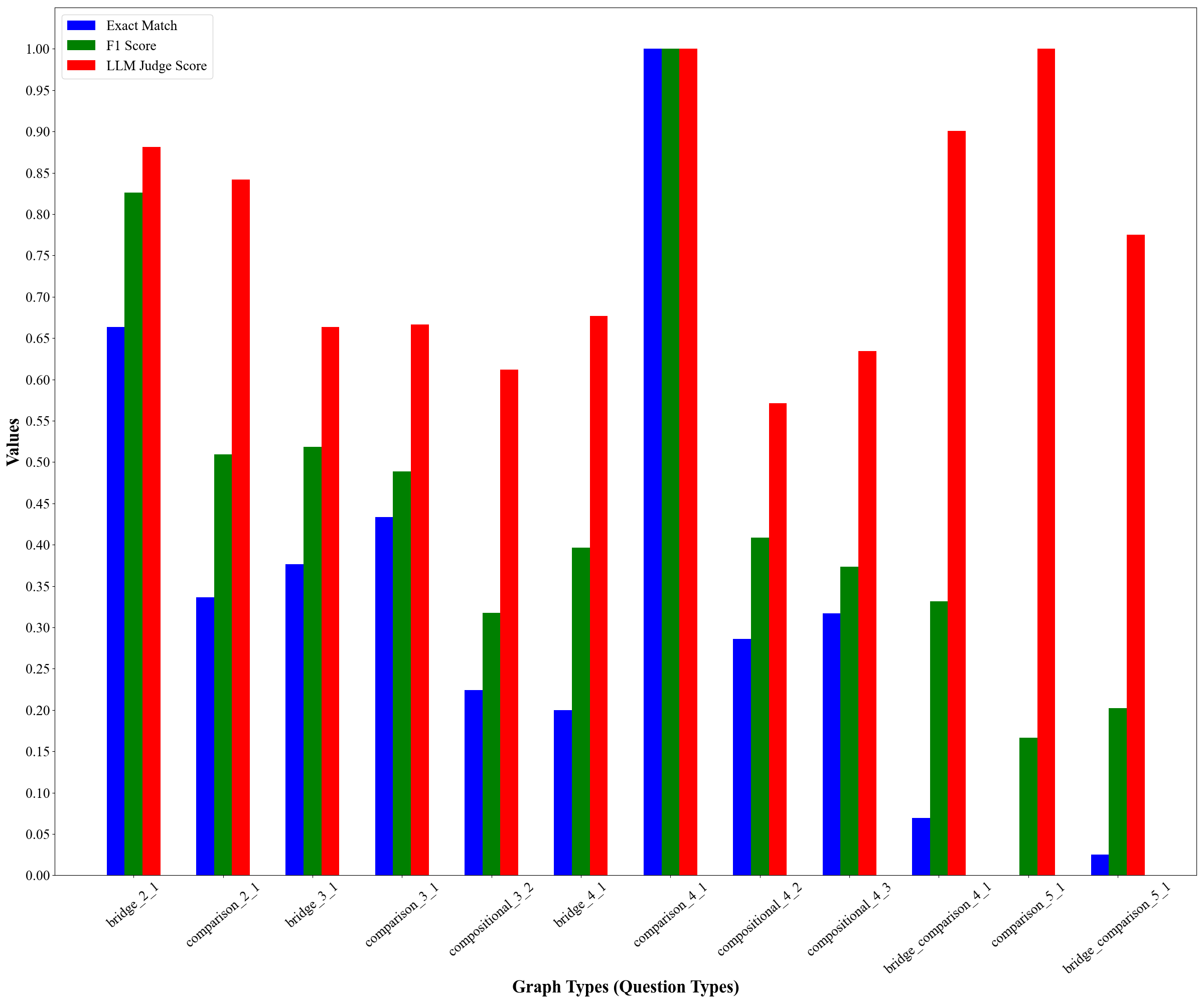}
    \caption{Llama3 Metrics - Positive Graph of Ground Truth Evidence}
    \label{fig:llama3_pos}
\end{figure}

\begin{figure}[htbp]
    \centering
    \includegraphics[width=1\linewidth]{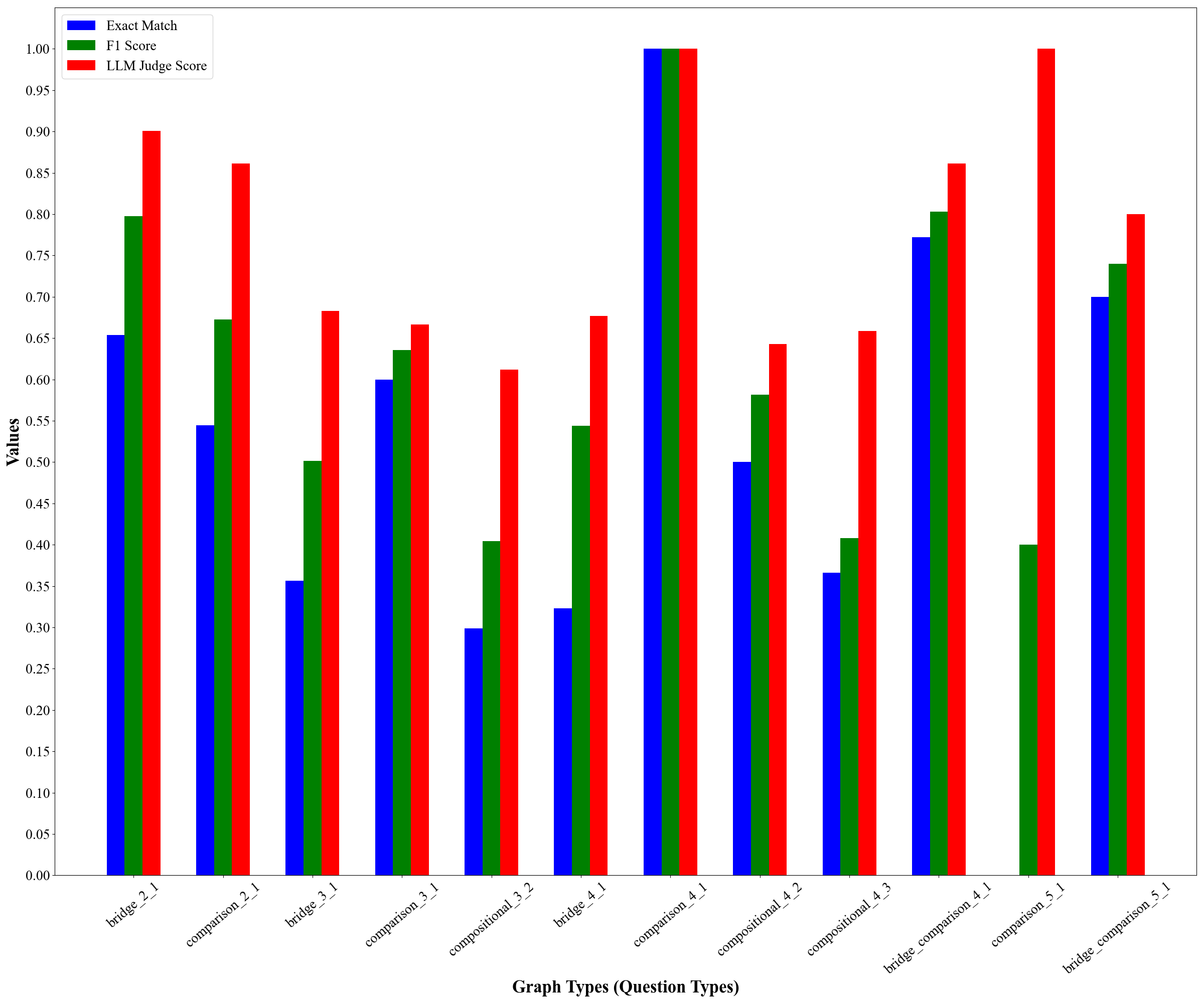}
    \caption{GPT-3.5 Metrics - Unstructured Ground Truth Evidence}
    \label{fig:3.5_no_structure}
\end{figure}

\begin{figure}[htbp]
    \centering
    \includegraphics[width=1\linewidth]{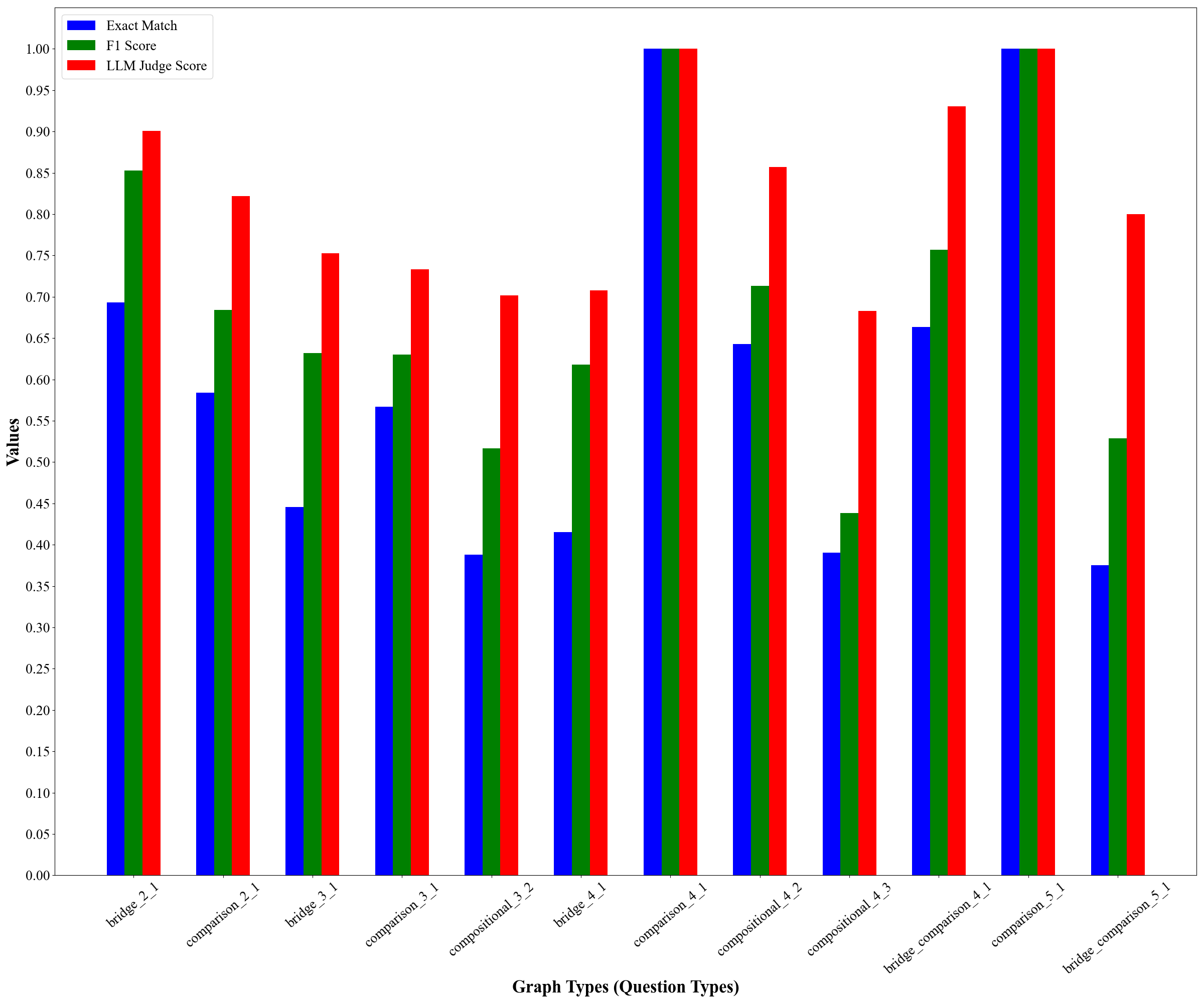}
    \caption{GPT4o-mini Metrics - Unstructured Ground Truth Evidence}
    \label{fig:4o-mini_no_structure}
\end{figure}

\begin{figure}[htbp]
    \centering
    \includegraphics[width=1\linewidth]{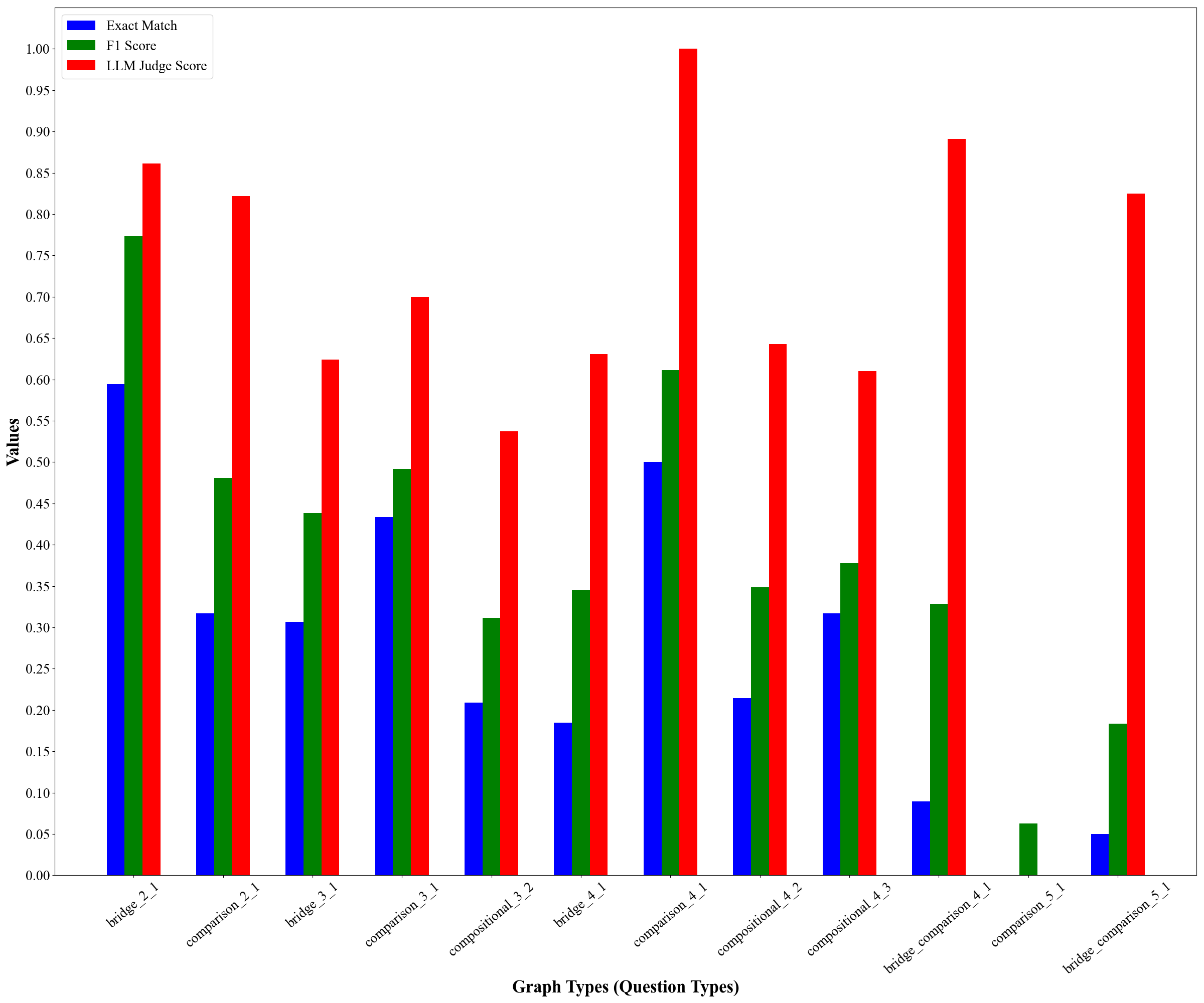}
    \caption{Llama3 Metrics - Unstructured Ground Truth Evidence}
    \label{fig:llama3_no_structure}
\end{figure}

\begin{figure}[htbp]
    \centering
    \includegraphics[width=1\linewidth]{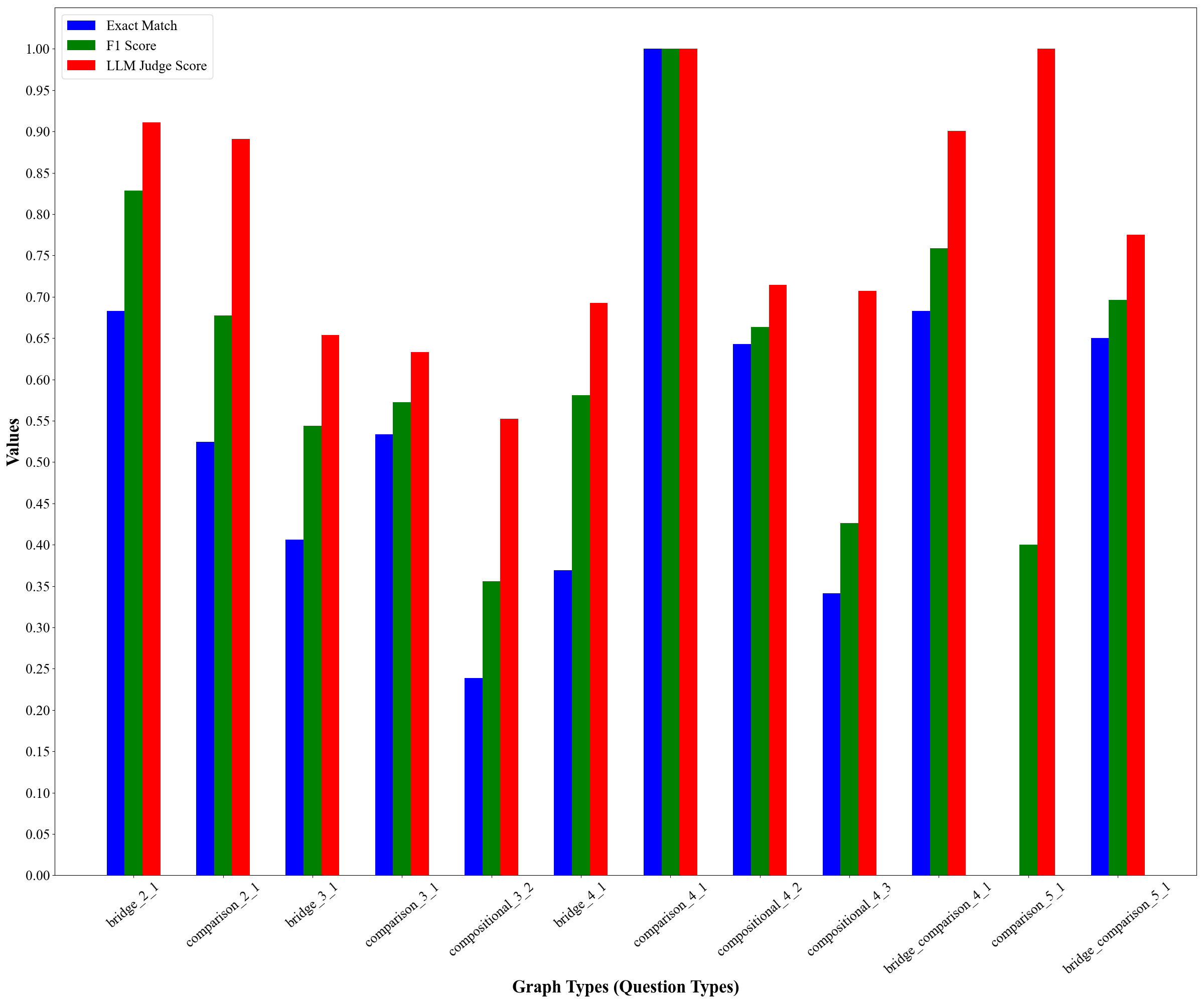}
    \caption{GPT-3.5 Metrics - Negative Graph of Ground Truth Evidence}
    \label{fig:3.5_neg}
\end{figure}

\begin{figure}[htbp]
    \centering
    \includegraphics[width=1\linewidth]{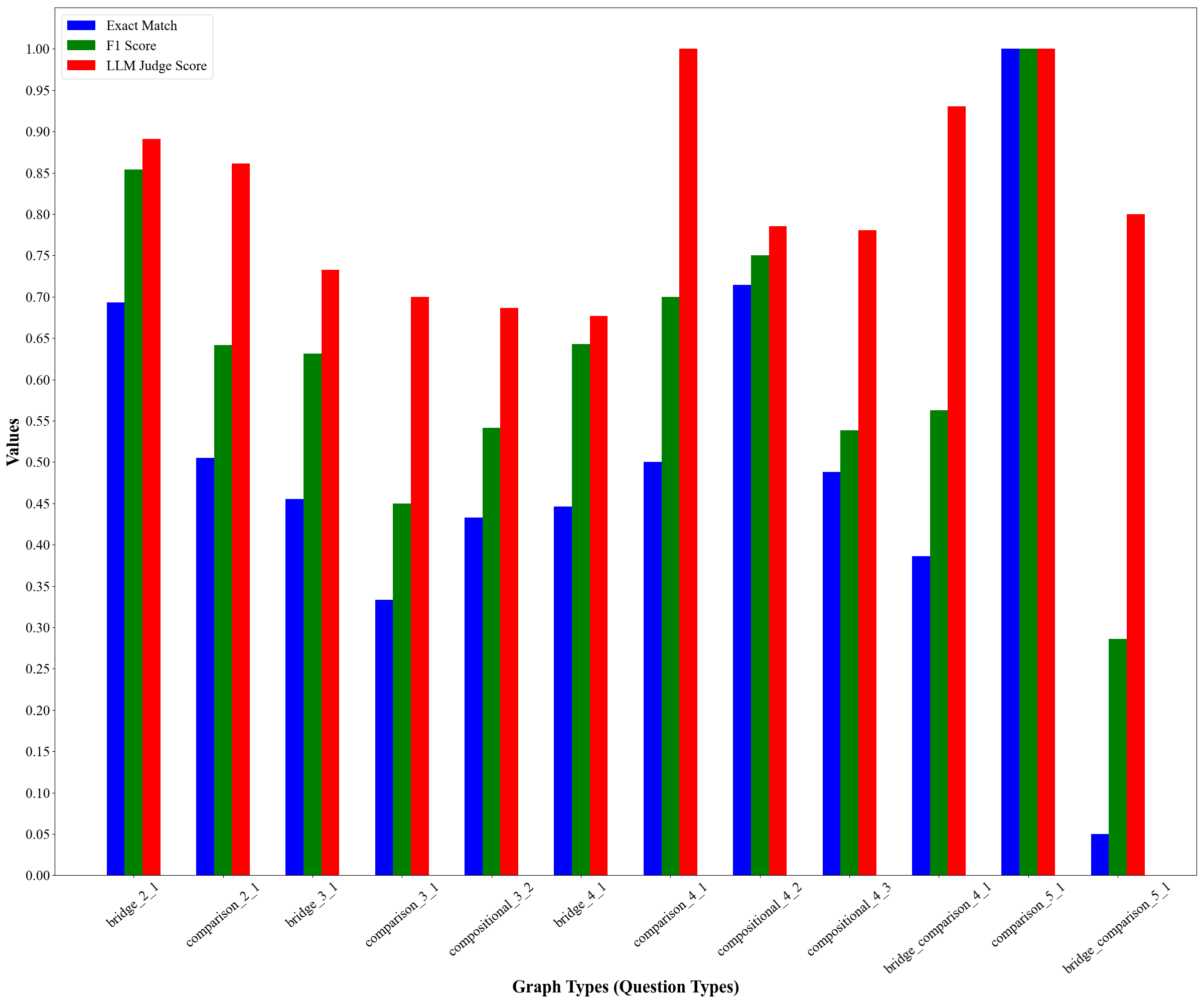}
    \caption{GPT4o-mini Metrics - Negative Graph of Ground Truth Evidence}
    \label{fig:4o-mini_neg}
\end{figure}

\begin{figure}[htbp]
    \centering
    \includegraphics[width=1\linewidth]{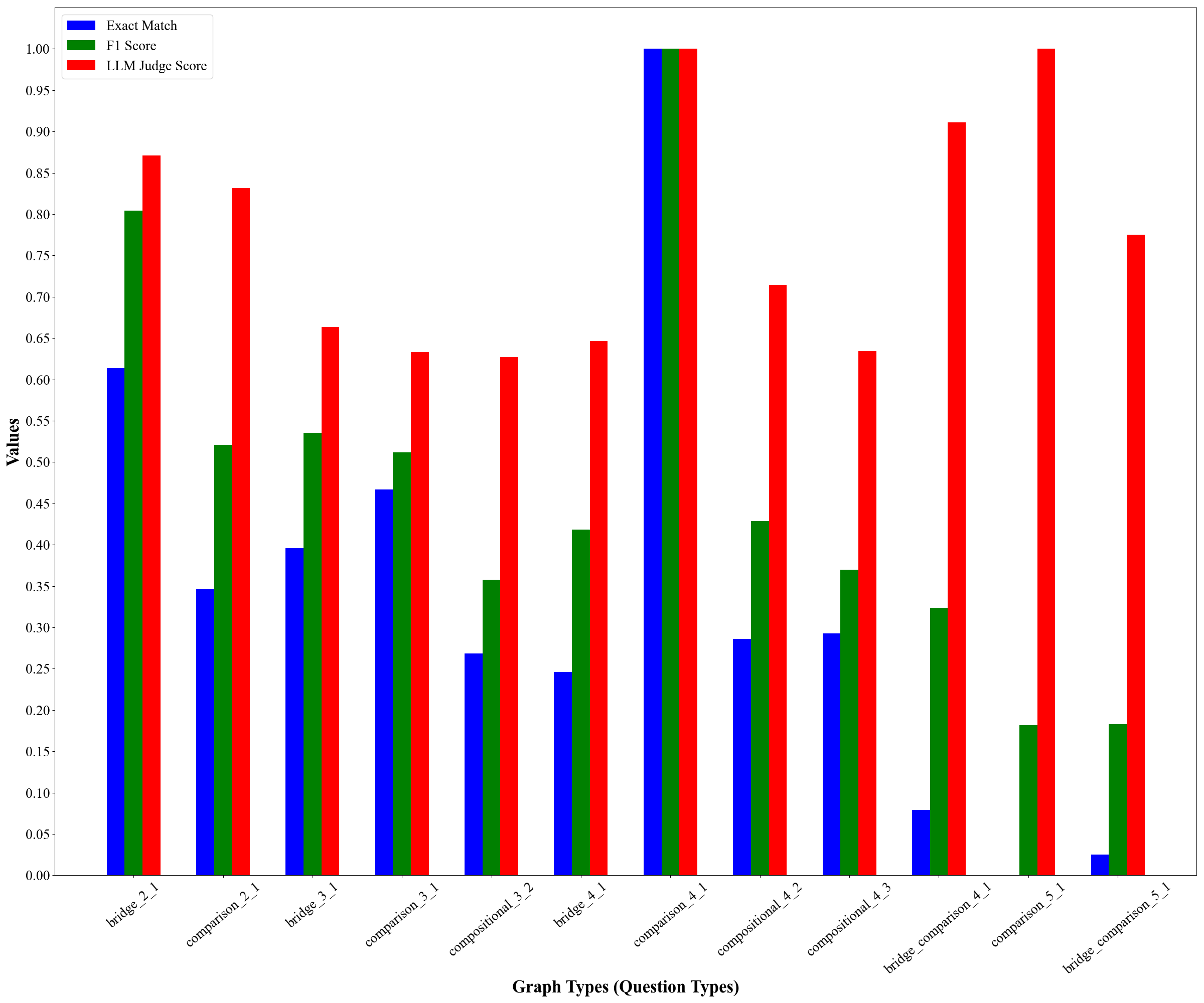}
    \caption{Llama3 Metrics - Negative Graph of Ground Truth Evidence}
    \label{fig:llama3_neg}
\end{figure}

\begin{figure}[htbp]
    \centering
    \includegraphics[width=1\linewidth]{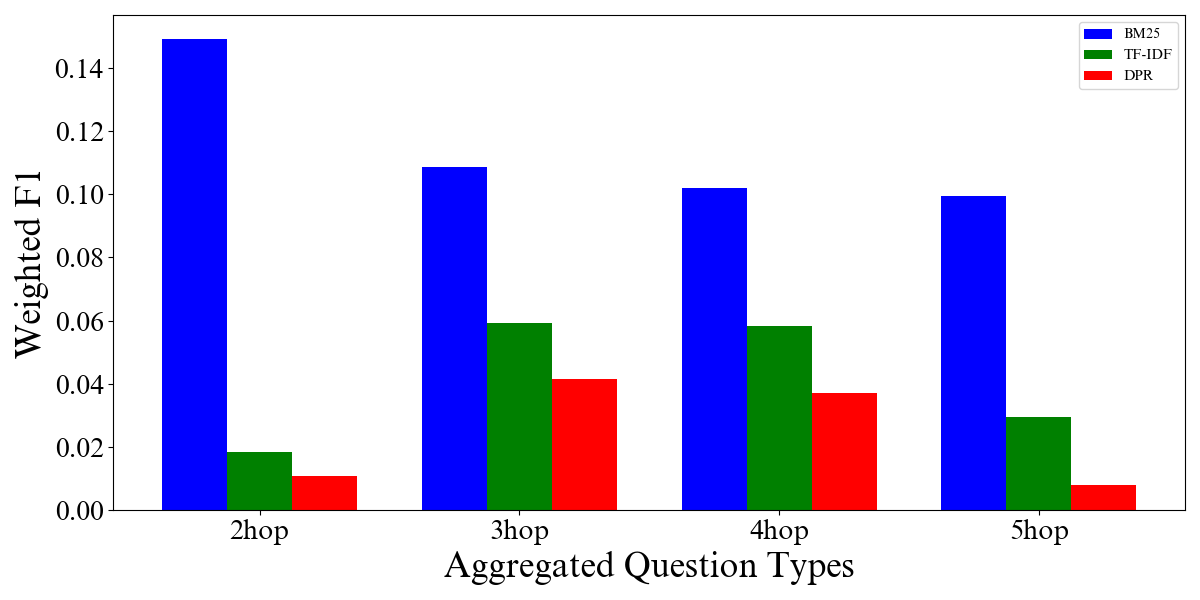}
    \caption{BM25, TFIDF, and DPR Weighted F1 Across Question Types}
    \label{fig:weighted-f1}
\end{figure}

\begin{figure}[htbp]
    \centering
    \includegraphics[width=1\linewidth]{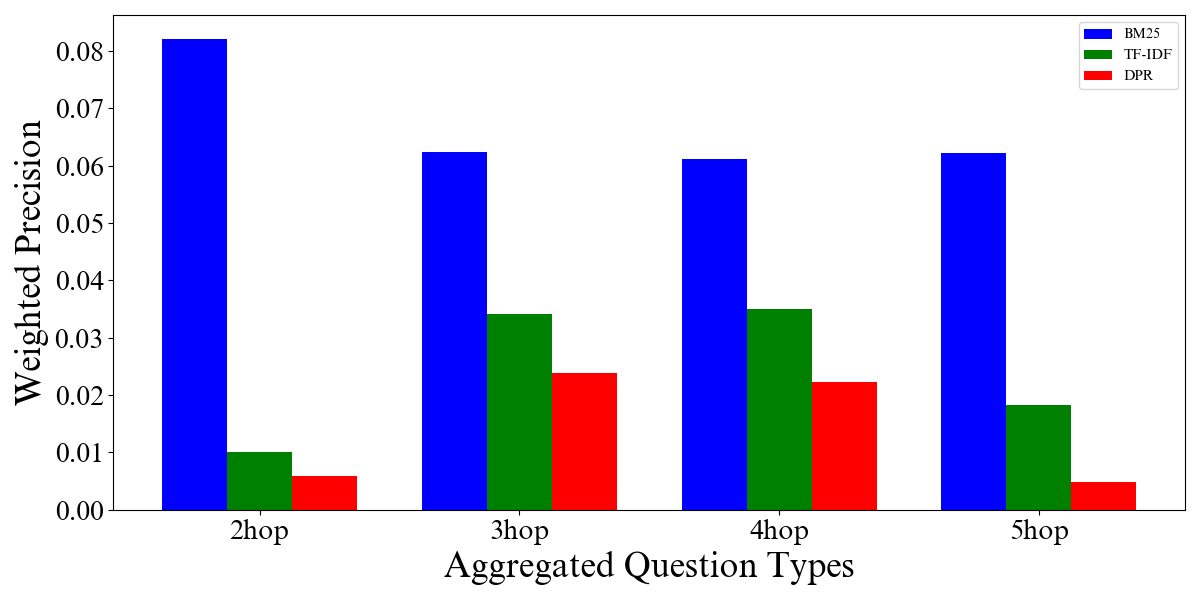}
    \caption{BM25, TFIDF, and DPR Weighted Precision Across Question Types}
    \label{fig:weighted-precision}
\end{figure}

\end{document}